%% file: nips_2018.tex
\documentclass{article}
\pdfoutput=1

\usepackage[preprint]{nips_2018}

\usepackage[utf8]{inputenc} 
\usepackage[T1]{fontenc}    
\usepackage{hyperref}       
\usepackage{url}            
\usepackage{booktabs}       
\usepackage{amsfonts}       
\usepackage{nicefrac}       
\usepackage{microtype}      

\usepackage{amsmath}
\usepackage{graphicx}

\graphicspath{{figures/}}

\DeclareMathOperator*{\argmin}{arg\,min}

\usepackage{calligra}
\DeclareMathAlphabet{\mathcalligra}{T1}{calligra}{m}{n}
\DeclareFontShape{T1}{calligra}{m}{n}{<->s*[2.2]callig15}{}

\newcommand{\thbthmat}[9]{\left(\begin{array}{ccc} #1 & #2 & #3 \\ #4 & #5 & #6 \\ #7 & #8 & #9 \end{array}\right)}

\title{Spiking Linear Dynamical Systems on Neuromorphic Hardware for Low-Power Brain-Machine Interfaces}

\author{
David G.~Clark\thanks{Biological Systems and Engineering Division, Lawrence Berkeley National Laboratory}\\
\texttt{dgc@lbl.gov}
\And
Jesse A.~Livezey\footnotemark[1]$\:\:^,$\thanks{Redwood Center for Theoretical Neuroscience, University of California, Berkeley}\\
\texttt{jlivezey@lbl.gov}
\AND 
Edward F.~Chang\thanks{Department of Neurological Surgery and Department of Physiology, University of California, San Francisco}$\:\:^,$\thanks{Center for Integrative Neuroscience, University of California, San Francisco}$\:\:^,$\thanks{UCSF Epilepsy Center, University of California, San Francisco}\\
\texttt{edward.chang@ucsf.edu}
\And 
Kristofer E.~Bouchard\footnotemark[1]$\:\:^,$\footnotemark[2]$\:\:^,$\thanks{Helen Wills Neuroscience Institute, University of California, Berkeley}$\:\:^,$\thanks{To whom correspondence should be addressed}\\
\texttt{kebouchard@lbl.gov}
}

\begin{document}

\newcommand{\tbtmat}[4]{\left(\begin{array}{cc} #1 & #2 \\ #3 & #4 \end{array}\right)}\newcommand{\tbomat}[2]{\left(\begin{array}{c} #1 \\ #2 \end{array}\right)}

\newcommand{\diag}[1]{\text{diag}\left(#1\right)}
\newcommand{\bo}[1]{\mathbf{#1}}
\newcommand{\bs}[1]{\boldsymbol{#1}}
\newcommand{\relu}[1]{\text{ReLU}\left( #1 \right)}

\maketitle

\begin{abstract}
Neuromorphic architectures achieve low-power operation by using many simple spiking neurons in lieu of traditional hardware. Here, we develop methods for precise linear computations in spiking neural networks and use these methods to map the evolution of a linear dynamical system (LDS) onto an existing neuromorphic chip: IBM's TrueNorth. We analytically characterize, and numerically validate, the discrepancy between the spiking LDS state sequence and that of its non-spiking counterpart. These analytical results shed light on the multiway tradeoff between time, space, energy, and accuracy in neuromorphic computation. To demonstrate the utility of our work, we implemented a neuromorphic Kalman filter (KF) and used it for offline decoding of human vocal pitch from neural data. The neuromorphic KF could be used for low-power filtering in domains beyond neuroscience, such as navigation or robotics.
\end{abstract}

\section{Introduction}

Neuromorphic computing \cite{mead} has seen a resurgence due to the potential use of neuromorphic architectures as a low-power computing framework. Many applications have focused on deep learning inference, which requires multilayer feedforward computations to be mapped onto spiking chips \cite{Esser201604850, rueckauer2017conversion}, possibly with novel training algorithms \cite{NIPS20155862, hunsberger2016training, zenke2017superspike}. However, little work has been done to map \textit{recurrent} computations onto neuromorphic chips, perform \textit{precise} linear computations using spikes, or obtain \textit{analytical descriptions} of the discrepancies between spiking computations and their real-valued counterparts. We addressed these open problems in the setting of linear dynamical systems (LDSs).

\par 
Several neuromorphic architectures have been recently introduced, ranging from flexible but energy-intensive to inflexible but energy-efficient \cite{schuman2017survey}. For instance, the SpiNNaker system consists of many ARM cores and is highly flexible since neurons are implemented at the software level, albeit energy-intensive (each core consumes $\sim$1 W) \cite{furber2014spinnaker}. IBM's TrueNorth, by contrast, is relatively inflexible since each core implements hardware for a small group of simple neurons, but is extremely energy-efficient (the whole chip consumes $\sim$70 mW) \cite{merolla2014million, carney2017neuromorphic}. To reap the rewards of energy-efficiency, we targeted TrueNorth, and addressed the challenges associated with mapping computation onto a tightly-constrained architecture. In particular, we developed a precise mapping of LDSs onto a spiking neural network which obeys TrueNorth's constraints.

\par
Brain-machine interfaces (BMIs) restore lost function by mapping neural activity to kinematic variables in real-time, however challenges remain in deploying such systems for everyday use. In particular, decoding algorithms are often computationally demanding and thus dissipate significant energy operating continuously. The Kalman filter (KF), the optimal Bayesian decoder under certain modeling assumptions, works well in practice and is widely used for BMIs \cite{kalman1960new, malik2011efficient, gilja2012high}. Previous work has explored using recurrent neural network-based decoders for BMIs with the eventual goal of neuromorphic deployment \cite{sussillo2012recurrent, dethier2013design}, however to our knowledge no groups have mapped a decoder onto an actual neuromorphic chip. As mentioned above, this is significantly more difficult than targeting an unconstrained abstract model. Our work, which targets TrueNorth specifically, therefore represents a major step in the progression from exploratory research to deployment.

\section{Methods}
\label{sec:methods}

TrueNorth is a fully-digital neuromorphic architecture created by IBM consisting of 4,096 cores joined in a square mesh network. The chip operates in discrete time, uses binary spikes, and has integer-valued parameters.  Because TrueNorth is fully digital, it may be simulated spike-for-spike on a regular computer. We therefore created a Python simulation of the spiking LDS for testing, which we validated against a version programmed in the chip's proprietary MATLAB API and deployed on a TrueNorth NS1e test board \cite{sawada2016truenorth}.

\par 
Many features of the TrueNorth neuron model are not relevant for our implementation. We therefore work in terms of a simplified neuron which is both subsumed by the TrueNorth model (e.g., it uses integer-valued parameters and binary spikes) and is sufficient for our implementation; in Section \ref{subsec:map}, we describe final modifications which make the spiking LDS compatible with TrueNorth's full set of constraints. Our neuron model has $d$ inputs, integer membrane potential $v_i \in \mathbb{Z}$ (where $i$ indexes discrete time), integer synaptic weights $\bs{\alpha} \in \mathbb{Z}^d$, and positive integer firing threshold $\beta \in \mathbb{Z}_+$. Let $\bo{x}_i \in \{0, 1\}^d$ describe the input spike pattern at time-step $i$. At $i \geq 1$, the neuron performs an integrate-and-fire update:
\begin{equation}
\begin{split}
&v_{i} \gets v_{i-1} + \bs{\alpha} \cdot \bo{x}_i  \\
&\text{if } v_i \geq \beta :\\
&\:\:\:\:\:\textbf{spike}\\
&\:\:\:\:\:v_i \gets v_i - \beta
\end{split}
\label{eq:neuron}
\end{equation}
Upon firing, this neuron changes its membrane potential by subtracting its firing threshold, imbuing the neuron with a ``memory'' which persists between firing events. This memory allows it to perform scalar multiplication in an unbiased fashion.

\par 
We developed a spiking version of a discrete-time, time-invariant, driven LDS of the form
\begin{equation}
\bo{x}_{t} = A \bo{x}_{t-1} + B \bo{u}_t, \:\:\: \bo{x}_{0} = 0
\end{equation}
where $A$ is the dynamics matrix, $B$ is the input matrix, and $\bo{x}_t$ and $\bo{u}_t$ are the state and input, respectively, at time $t$. To assess the behavior of the spiking LDS, we computed the residuals $\bo{r}_t$ between the state sequences yielded by the spiking LDS and its non-spiking counterpart under equivalent inputs. We then divided the residuals by a factor related to the spike coding scheme to obtain normalized residuals $\tilde{\bo{r}}_t$. Based on the normalized residuals we computed the sample covariance matrix and mean squared error (MSE), given by
\begin{equation}
\Sigma_\text{sample} = \frac{1}{T} \sum_{t=1}^T \tilde{\bo{r}}_t \tilde{\bo{r}}_t^T,
\:\:\:
\text{MSE}_\text{sample} = \text{tr}\left( \Sigma_\text{sample} \right).
\label{eq:samplecov}
\end{equation}
The calculation of $\Sigma_\text{sample}$ does not include mean-subtraction since our analysis predicts zero-mean residuals. Our main analytical contribution is to derive the covariance matrix of the residuals given weak statistical assumptions about the inputs, thus providing predictions for the quantities in Eq. \ref{eq:samplecov}. To our knowledge, we are the first to derive such a result. For the purposes of our numerical experiments, which validate these predictions, we used LDSs with random system matrices and sinusoidal inputs.

\par
To demonstrate our work's utility, we decoded human vocal pitch from simultaneously recorded electrocorticography (ECoG) neural activity using a spiking version of the KF. To generate a training set, a human subject repeatedly uttered the sentence ``I never said she stole my money'' while a microphone recorded their vocal pitch and an ECoG array placed on the subject's cortical surface recorded their neural activity \cite{dichter2018control}. The raw ECoG voltages were common average (median) referenced and converted to the time-frequency domain via the Hilbert transform. The $z$-scored analytic amplitude in the high-gamma band (75-150 Hz) was then extracted. Both the high-gamma and vocal pitch data were segmented into 38 trials, each trial encapsulating a single utterance of the sentence, and downsampled by averaging over non-overlapping windows to obtain 40 Hz data. We considered only the 73 electrodes covering the ventral sensorimotor cortex of the subject \cite{bouchard2013functional}. The experimental protocol was approved by the Human Research Protection Program at the University of California, San Francisco.

\section{Results}
\label{subsec:results}

\subsection{Representational tradeoffs \& nonnegativity}
\label{subsec:nonneg}

Mapping LDSs onto a spiking architecture required representing numerical values using binary spikes. To do this, we divided discretized time into non-overlapping ``frames'' of length $\ell$ and encoded values as spike counts over individual frames. Because this scheme may represent $\ell$ distinct values per frame, and each time-step requires some amount of physical time (1 ms for TrueNorth), \textit{a tradeoff is induced between the precision of the spike code, which is related to computational accuracy, and latency}. To mitigate this, we employed ``$p$-dimensional'' spike trains, meaning that values were encoded by the total spike counts of populations of $p$ neurons over frames. This scheme may represent $p\ell$ distinct values per frame, allowing us to increase the precision of the spike code, and thus the accuracy of the computation, at the cost of space (neuronal footprint) rather than time (latency). We first describe the spiking LDS for $p=1$ and generalize to $p > 1$ in Section \ref{subsec:pdim}.

\par
When encoding values as spike counts, \textit{there is no natural way to represent negative values} -- for example, a spike has no sign bit. What's more, \textit{the nonnegative character of spikes breaks the symmetry between addition and subtraction} in the following sense. For a neuron to add two input spike trains, it needs merely to merge them together. This is an error-free operation independent of the timing of the input spikes. However, for a neuron to subtract one spike train from another (for example, using two equal-and-opposite ingoing synaptic weights), the input spikes must be temporally aligned for the result to be error-free since a spike in the negative channel cannot cancel out an earlier spike in the positive channel. Developing spiking versions of general LDSs with mixed-sign system matrices, inputs, and states, required us to overcome these challenges.

\subsection{Nonnegative matrix-vector multiplication with spikes}
\label{subsec:nnmv}

Given a single input, the neuron model of Eq. \ref{eq:neuron} performs scalar multiplication by $\frac{\alpha}{\beta}$ (see Appendix \ref{subsec:appdxscalarmult}). Using many such neurons, one may form a circuit which maps $n$ input spike trains to $m$ output spike trains so as to perform matrix-vector multiplication by a nonnegative, element-wise rational matrix $W \in \mathbb{Q}^+_{m \times n}$. Note that while the neuron model allows for negative synaptic weights, we nonetheless demand the nonnegativity of $W$ so that we may leverage error-free spiking addition (discussed in Section \ref{subsec:nonneg}). For each $(i,j)$, we instantiated a multiplication neuron with ingoing synaptic weight $\alpha_{ij} \in \mathbb{Z}_+$ and firing threshold $\beta_{ij} \in \mathbb{Z}_+$ such that $w_{ij} = \frac{\alpha_{ij}}{\beta_{ij}}$. Each input component was routed to the multiplication neurons for the corresponding column of $W$. We then created $m$ copies of an $n$-way addition neuron, configured by setting both the firing threshold and ingoing synaptic weights to 1, to perform the row sums (Appendix Fig. \ref{fig:linearcircuits}A). Letting $\bo{n}^\text{in}_t \in \mathbb{Z}_+^n$ and $\bo{n}_t^\text{out} \in \mathbb{Z}_+^m$ describe the input and output spike counts of the matrix-vector multiplication circuit at frame $t$, we may represent the computation performed by the circuit as $\bo{n}^\text{out}_t = W\bo{n}^\text{in}_t + \bs{\epsilon}_t$, where $\bs{\epsilon}_t$ is the error at frame $t$ due to spiking computation. Since the row sums are error-free, each component of $\bs{\epsilon}_t$ is the sum of the errors due to the multiplication neurons for the corresponding row of $W$. Under a weak statistical assumption about the input spikes (see Appendix \ref{subsec:appdxmatmult}), $\bs{\epsilon}_t$ satisfies
\begin{equation}
\bo{E}\left[ \bs{\epsilon}_t \right] = 0,
\:\:\: \text{Cov}\left( \bs{\epsilon}_{t + \Delta t}, \bs{\epsilon}_{t} \right) = \frac{n}{6} f(\Delta t)I_m 
\label{eq:vecstats}
\end{equation}
where
\begin{equation}
f\left(\Delta t \right) = \begin{cases}
  1 & |\Delta t| = 0\\
 -\frac{1}{2} & |\Delta t| = 1\\
 0 & |\Delta t| \geq 2.
 \end{cases}
 \label{eq:f}
\end{equation}
The factor of $n$ in in the covariance of Eq. \ref{eq:vecstats} reflects the fact that each sum includes contributions from $n$ multiplication neurons, since $W$ has width $n$. Meanwhile, the negative autocovariance between adjacent frames expressed in Eq. \ref{eq:f} reflects the intuition that if a neuron \textit{overshoots} the unbiased result at some frame, then its membrane potential is likely small at the end of the frame, so it is likely to \textit{undershoot} the unbiased result at the subsequent frame, and vice versa. Because the error variance does not scale with $\ell$, the computation may be made arbitrarily precise by increasing $\ell$.

\subsection{Spiking linear dynamical systems}

Due to its central nonnegativity assumption, the matrix-vector multiplication circuit of Section \ref{subsec:nnmv} may not trivially be adapted to create a spiking version of a general LDS. Recall that a LDS is asymptotically stable if and only if its dynamics matrix $A$ satisfies $\rho(A) < 1$, where $\rho(A)$ is the spectral radius of $A$, its largest eigenvalue in absolute value. While any asymptotically stable LDS with bounded inputs may be transformed so as to have nonnegative inputs and states by applying affine linear transformations to these variables, we may not transform away nonnegative elements of the dynamics matrix in the same way. In particular, under any affine linear transformations of the inputs and states, the transformed dynamics matrix is similar to the original one, however \textit{there exist matrices to which no nonnegative matrix is similar} (for example, any $A$ with $\text{tr}(A) < 0$). To handle general LDSs, we therefore transformed from a given LDS to another LDS of twice the size with nonnegative matrices, inputs, and states, and recovered from the transformed system's state sequence that of the original. To implement the transformed system with spikes, we adapted the techniques of Section \ref{subsec:nnmv}. Specifically, we instantiated a set of multiplication neurons for each of the system matrices, and formed recurrent connections from the outputs of the addition neurons, configured to perform sums over the rows of both matrices, back to the state inputs. These recurrent connections were imbued with a delay equal to the frame length $\ell$ minus the minimum number of time-steps required for a spike to propagate through the multiplication and addition stages (Appendix Fig. \ref{fig:linearcircuits}B).

\par 
Let the original LDS be given by $\bo{x}_t = A\bo{x}_{t-1} + B\bo{u}_t$, where $A \in Q_{m \times m}$, $B \in Q_{m \times n}$, $\bo{u}_t \in \{-\ell, \ldots, \ell\}$, and $\bo{x}_t \in [-\ell, \ell]$. The domains of the input and state simplify conversion to/from spikes. The spiking system, which has input dimension $2n$ and state dimension $2m$, is described by
\begin{equation}
\begin{split}
\tbomat{ \bo{n}^\text{state,+}_t }{\bo{n}^\text{state,-}_t}
&= \tbtmat{\relu{A}}{\relu{-A}}{\relu{-A}}{\relu{A}} \tbomat{ \bo{n}^\text{state,+}_{t-1} }{\bo{n}^\text{state,-}_{t-1}}\\
&+ \tbtmat{\relu{B}}{\relu{-B}}{\relu{-B}}{\relu{B}} \tbomat{ \bo{n}^\text{in,+}_{t} }{\bo{n}^\text{in,-}_{t}}
+ \tbomat{ \bs{\epsilon}^+_t }{ \bs{\epsilon}^-_t }
\end{split}
\label{eq:biglds}
\end{equation}
where $\text{ReLU}(x) = \text{max}(0, x)$ and $\bs{\epsilon}^+_t$ and $\bs{\epsilon}^-_t$ are the errors due to spiking computation at frame $t$. The spiking LDS inputs are given in terms of those of the original LDS according to $\bo{n}^{\text{in},+}_t = \relu{ \bo{u}_t }$ and $\bo{n}^{\text{in},-}_t = \relu{- \bo{u}_t }$. If we set $\bs{\epsilon}^+_t = \bs{\epsilon}^-_t = 0$ in Eq. \ref{eq:biglds}, then the state sequence of the original LDS may be perfectly recovered from that of the spiking LDS according to $\bo{x}_t = \bo{n}^{\text{state},+}_t - \bo{n}^{\text{state},-}_t$. The proof leverages identities satisfied by the twice-as-large system matrices of Eq. \ref{eq:biglds}, denoted $\tilde{A}$ and $\tilde{B}$. In particular, defining the operators $[\cdot]^+$ and $[\cdot]^-$ to pick out the top and bottom halves, respectively, of a column vector with even dimension, then for any $\bo{v} \in \mathbb{R}^{2m}$, $\bo{w} \in \mathbb{R}^{2n}$, and $k \geq 0$,
\begin{equation}
\left[\tilde{A}^k\bo{v} \right]^+ - \left[\tilde{A}^k\bo{v} \right]^- = A^k\left(\left[\bo{v}\right]^+ - \left[\bo{v}\right]^- \right), \:\:\: \left[\tilde{B}\bo{w} \right]^+ - \left[\tilde{B}\bo{w} \right]^- = B\left(\left[\bo{w}\right]^+ - \left[\bo{w}\right]^- \right).
\label{eq:identities}
\end{equation}
We implemented the spiking system of Eq. \ref{eq:biglds} on TrueNorth, read out $\bo{n}^{\text{state},+}_t$ and  $\bo{n}^{\text{state},-}_t$ at each frame, and performed the subtraction step off-chip to recover the state sequence of the original LDS. Under this implementation, the residual at frame $t$ between the spiking and non-spiking state sequences is
\begin{equation}
\bo{r}_t = \left[ \bo{n}^{\text{state},+}_t - \bo{n}^{\text{state},-}_t \right] - \bo{x}_t = \sum_{k=0}^{t-1} A^k \left( \bs{\epsilon}_{t-k}^+ - \bs{\epsilon}_{t-k}^- \right).
\label{eq:accumulateddiff}
\end{equation}
Assuming that $\rho(A) < 1$, this error does not diverge. Under a weak statistical assumption about the input spikes (see Appendix \ref{subsec:appdxlds}), the difference $\Delta \bs{\epsilon}_t = \bs{\epsilon}_{t}^+ - \bs{\epsilon}_{t}^-$ satisfies
\begin{equation}
\bo{E}\left[ \Delta \bs{\epsilon}_t  \right] = 0, \:\:\:
\text{Cov}\left( \Delta \bs{\epsilon}_{t + \Delta t} , \Delta \bs{\epsilon}_{t}  \right) = 
\frac{2m + n}{6}f(\Delta t) I_m
\label{eq:statsmixedldsmaintext}
\end{equation}
where $2m + n$ is the total number of scalar multiplication results added to form $n^{\text{state},+}_{t,i}$ and $n^{\text{state},-}_{t,i}$; the input dimension $n$ does not come with a factor of $2$ since for each component of $\bo{n}^{\text{in},+}_t$ which is nonzero, the corresponding component of $\bo{n}^{\text{in},-}_t$ is zero, and vice versa. It follows from substitution of Eq. \ref{eq:statsmixedldsmaintext} into Eq. \ref{eq:accumulateddiff} that
\begin{equation}
\bo{E}\left[ \bo{r}_t \right] = 0,\:\:\: \text{Cov}\left(\bo{r}_{t \rightarrow \infty}\right) = \frac{2m + n}{6}\text{sym}\left( (I-A)\sum_{k=0}^\infty A^k \left( A^k \right)^T \right)
\label{eq:mixedcovinf}
\end{equation}
where $\text{sym}\left(X\right) = \frac{1}{2}\left( X + X^T\right)$, and
\begin{equation}
\text{Cov}\left(\bo{r}_{t + \Delta t}, \bo{r}_t\right) = A^{\Delta t}\text{Cov}\left(\bo{r}_{t}\right) \:\:\: \left(\Delta t \geq 0, \: t \rightarrow \infty \right).
\label{eq:auto}
\end{equation}
Due to Eq. \ref{eq:auto}, the residuals have serial correlations with timescale $\tau_A \sim 1 / (\log 1 / \rho(A))$. Thus, even though the spiking LDS is an unbiased estimator of its non-spiking counterpart in the sense that $\bo{E}\left[ \bo{r}_t \right] = 0$, the residuals nonetheless exhibit structure across time. See Appendix \ref{subsec:appdxidproof}-\ref{subsec:appdxserialcorr} for details.

\subsection{Spike overflow \& stability of the spiking system}
\label{subsec:overflow}

The spiking LDS of Eq. \ref{eq:biglds} might exhibit a ``spike overflow'' effect. In this case, excessive spikes are routed from multiplication neurons to an addition neuron such that the addition neuron, capable of firing only once at each time-step, is forced to fire some spikes during the subsequent frame which were intended to be fired during the current frame. To mitigate this effect, note that the state sequence recovered via subtraction is invariant under the subtraction of any constant from both $n^{\text{state},+}_{t,i}$ and $n^{\text{state},-}_{t,i}$. This suggests that we introduce circuitry which, in addition to summing the spikes for $n^{\text{state},+}_{t,i}$ and $n^{\text{state},-}_{t,i}$, also cancels out some number of spikes present in the contributions for both. The cancellation circuit (Fig. \ref{fig:crossbar}A) achieves this behavior by maintaining the invariant that the membrane potentials of its two neurons are equal-and-opposite.

\par 
Under general conditions of the original LDS, control of overflow is a necessity, since even if the dynamics matrix $A$ of the original LDS satisfies $\rho(A) < 1$, the transformed dynamics matrix $\tilde{A}$ of Eq. \ref{eq:biglds} might not satisfy $\rho(\tilde{A}) < 1$. A necessary and sufficient condition for $\rho(\tilde{A}) < 1$ is $\rho(\text{abs}(A)) < 1$, where $\text{abs}(A)$ is the element-wise absolute value of $A$ (see Appendix \ref{subsec:appdxnecsuf}). Thus, if $\rho(\text{abs}(A)) \geq 1$, then the spiking system is not asymptotically stable, and without the cancellation circuitry, the network saturates with spikes and is prevented from performing the desired computation. When $\rho(\text{abs}(A)) \geq 1$, even with the cancellation circuitry in place, the network exhibits ``spontaneous'' spiking activity after input spikes to the network have ceased.

\subsection{Generalizing to \textbf{\textit{p}}-dimensional spike trains}
\label{subsec:pdim}

The primary challenge in generalizing the spiking LDS for $p$-dimensional spike trains is devising a $p$-dimensional version of scalar multiplication with error variance which does not scale with $p$ (as it would if a multiplication neuron was simply replicated $p$ times). Given a multiplier $\frac{\alpha}{\beta}$, with $\alpha, \beta \in \mathbb{Z}_+$, we constructed such a circuit by instantiating $p$ neurons with linearly ascending firing thresholds $\beta, 2\beta, \ldots, p\beta$ and full recurrent connectivity: each pair is joined by a symmetric inhibitory connection with weight $-\beta$, while each neuron has an excitatory self-connection with weight $(i-1)\beta$, where $i \in \{1, \ldots, p\}$ indexes the neurons. Each successive neuron computes another digit in a unary representation of the result. Input spikes are routed to the population in a fully-connected fashion with synaptic weight $\alpha$ (Fig. \ref{fig:crossbar}B). Note that the multiplier denominator $\beta$, which plays the role of the firing threshold in the single-neuron case, also serves as a synaptic weight \textit{with a negative sign} in the $p$-dimensional case. This circuit maintains the invariant that the membrane potentials of all $p$ neurons are equal at all time-steps. One can show that the the $p$-dimensional circuit emulates in a single time-step the behavior of a single multiplication neuron over a frame of length $p$, where the shared membrane potential of the $p$ neurons corresponds to that of the single neuron. Thus, we may vary the representational capacity $p \ell$ of the code by modulating either $p$ or $\ell$ while the error statistics remain fixed as in the single-neuron case. This means that the computation may be made arbitrarily precise by increasing $p\ell$. When generalizing the addition and cancellation circuits for $p$-dimensional spike trains, one has the choice of either replicating the $p=1$ circuits $p$ times, or adapting the solution for the multiplication circuit described here. We chose the latter because it is less prone to spike overflow.

\subsection{Mapping onto TrueNorth}
\label{subsec:map}

\begin{figure}
  \centering
  \includegraphics[width=5.5in]{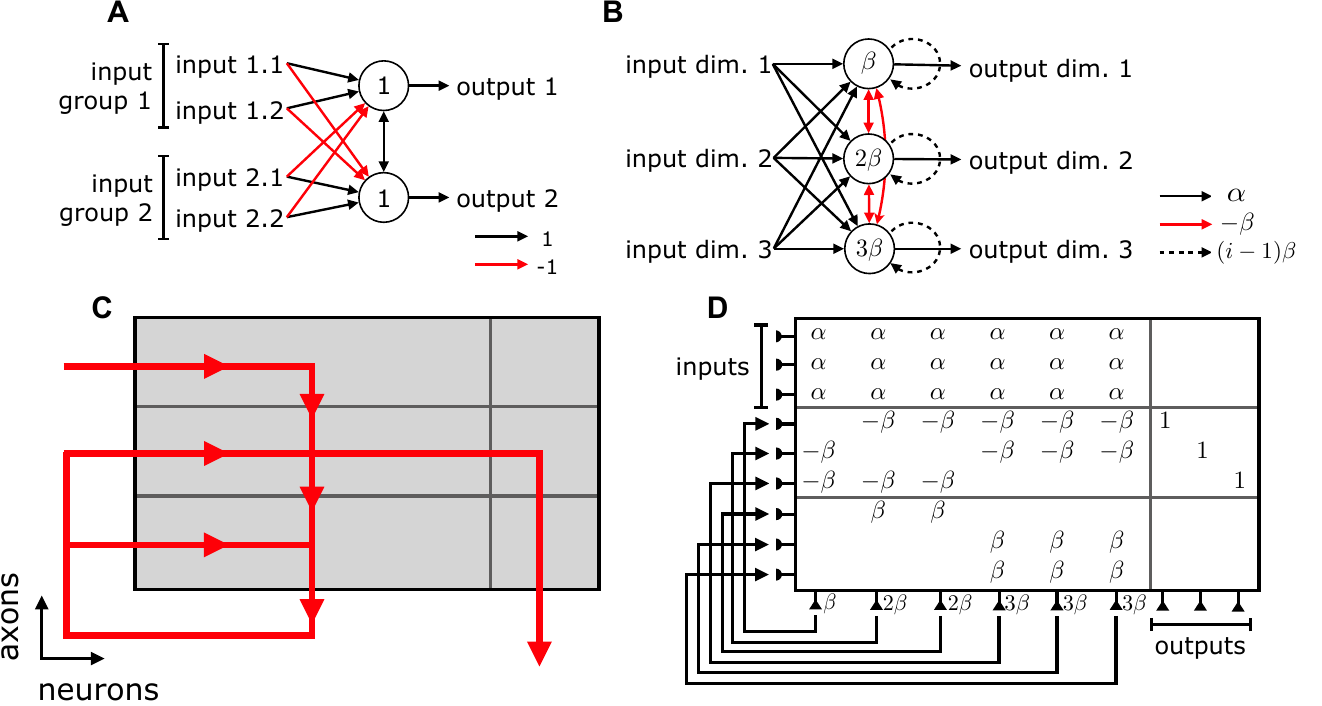}
  \caption{\textbf{(A)} Cancellation circuit. \textbf{(B)} $p$-dimensional multiplication circuit for $p = 3$. \textbf{(C)} Sketch of the flow of spikes in the $p$-dimensional multiplication crossbar. \textbf{(D)} $p$-dimensional multiplication crossbar for $p = 3$, i.e. the TrueNorth version of (B).}
  \label{fig:crossbar}
\end{figure}

Each TrueNorth core has 256 ``axons,'' which receive spikes, and 256 ``neurons,'' which fire spikes. Each axon routes spikes to up to all of the neurons on the same core via a fully-connected crossbar containing synaptic weights. Each neuron, in turn, routes spikes to exactly one axon on any core. A core's crossbar weight matrix is determined via the following procedure. First, each axon $i$ is assigned a label $G_i \in \{0, 1, 2, 3\}$, called an axon type. Then, each neuron $j$ is assigned four numbers $s^0_j, s^1_j, s^2_j, s^3_j \in \{-255, \ldots, 255\}$. Finally, a binary matrix $B \in \{0, 1\}^{256 \times 256}$ is configured. The weight between axon $i$ and neuron $j$ is then given by $w_{ij} = b_{ij} s^{G_i}_j$. Mapping the circuits for multiplication, addition, and cancellation onto TrueNorth is nontrivial due to the division of TrueNorth into finite crossbars, connectivity constraints, and the parameterization of synaptic weights.

\par 
In particular, the $p$-dimensional multiplication circuit (see Section \ref{subsec:pdim}) is incompatible with the required parameterization of synaptic weights for $p > 3$ due to the ascending excitatory self-connections, which demand $p$ distinct values of axon types, plus one more for the feedforward input connections. We therefore applied a series of transformations which render the circuit compatible with TrueNorth while increasing required number of neurons and axons to $n_\text{mult}(p) = \frac{1}{2}p^2 + \frac{3}{2}p$ (Fig. \ref{fig:crossbar}C-D). Since each multiplication circuit must fit on a single core, $n_\text{mult}(p) \leq 256$, so $p \leq 21$. Additionally, the multiplier numerator and denominator are constrained to be in $\{0, 1, \ldots, 255\}$, so the system matrices of the spiking LDS are perturbed versions of the original ones due the rational approximation error. However, the resulting perturbation to the state sequence was negligible for the purposes of our experiments. Finite core size limits the largest $p$-dimensional addition or cancellation circuit which fits on a single core, so we constructed $(m+n)$-way addition/cancellation circuits out of multiple $k$-way addition/cancellation circuits for $k < m+n$ in a way that minimized neuronal footprint. Finally, dedicated splitter crossbars were employed to handle TrueNorth's connectivity constraints. See Appendix \ref{sec:appdxtnmap} for details.

\subsection{Validation of error model}
\label{subsec:validatemodel}

\begin{figure}
  \centering
  \includegraphics[width=5.5in]{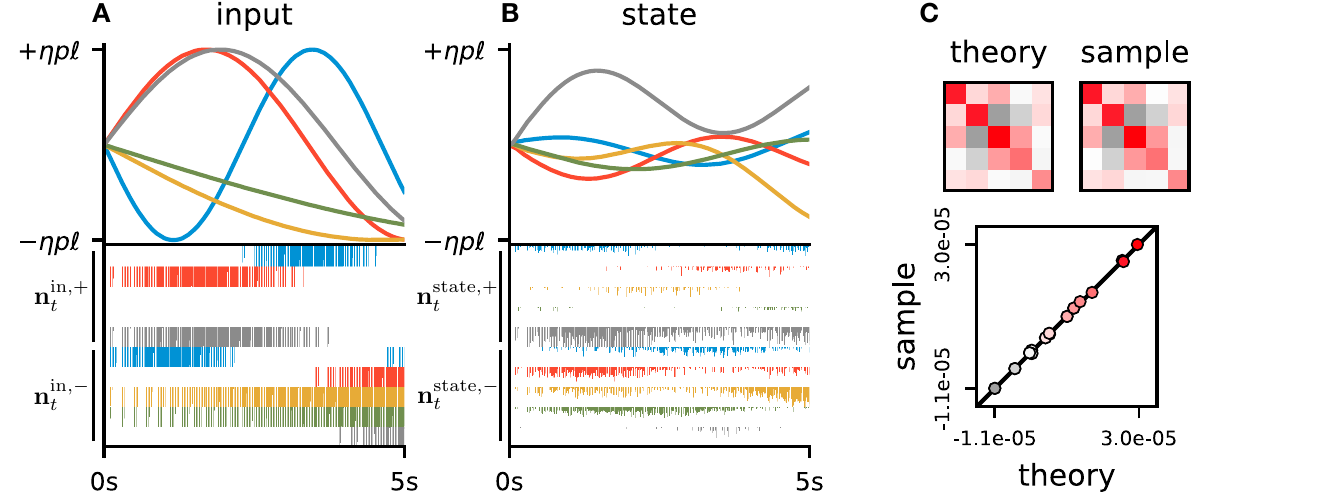}
  \caption{Validating the analytical covariance using a randomly generated LDS with 5-dimensional inputs and states. \textbf{(A)} First 5 seconds of the inputs (top) and equivalent spiking inputs (bottom). \textbf{(B)} First 5 seconds of the state sequence (top) and spiking state sequence (bottom). \textbf{(C)} Comparison of the theoretical and sample covariance matrices of the normalized residuals. }
  \label{fig:mainexperiment}
\end{figure}

To compare the observed and predicted covariance matrices of the normalized residuals between the spiking and non-spiking LDS state sequences, we implemented a randomly generated LDS on TrueNorth (see Appendix \ref{sec:appdxldsgen}). The inputs and states of the non-spiking system were normalized to $[-\eta p \ell, \eta p \ell]$, where $0 < \eta \leq 1$ controlled the maximum saturation of the input spike trains (we used $\eta = 0.9, p = 21$, and $\ell = 25$), and the inputs were integer-quantized (Fig. \ref{fig:mainexperiment}A-B, top panels). We drove the spiking LDS with the corresponding input spikes and measured the system's output spikes (Fig. \ref{fig:mainexperiment}A-B, bottom pannels). We then computed the normalized residuals as $\tilde{\bo{r}}_t = \frac{1}{\eta p \ell}\bo{r}_t$. The sample covariance matrix of the normalized residuals (Eq. \ref{eq:samplecov}) was then compared to the theoretical prediction, obtained by dividing the covariance of Eq. \ref{eq:mixedcovinf} by $\left(\eta p \ell\right)^2$:
\begin{equation}
\Sigma_\text{theory} = \frac{2m + n}{6\eta^2 p^2 \ell^2}\text{sym}\left((I-A)\sum_{k=0}^\infty A^k \left(A^k\right)^T \right).
\label{eq:errmodel}
\end{equation}
We observed good agreement between theory and experiment (Fig. \ref{fig:mainexperiment}C). Note that Eq. \ref{eq:errmodel} makes manifest various tradeoffs in neuromorphic computing. For fixed $m$ and $n$, the number of neurons in the spiking LDS is proportional to $p$, and the power consumption of TrueNorth is to a good approximation proportional to the number of active neurons \cite{merolla2014million}. Since $\ell$ is the length in milliseconds of each frame, the factor $p\ell$ is proportional to the the energy-per-frame of the computation. Therefore, computational accuracy (measured by inverse normalized residual variance) depends only on the energy-per-frame, without preference about the division of energy expenditure between space ($p$) and time ($\ell$). Meanwhile, the inverse scaling of computational accuracy with the system dimensions $m$ and $n$ reflects the distributed nature of computation.

\par
\begin{figure}
  \centering
  \includegraphics[width=5.5in]{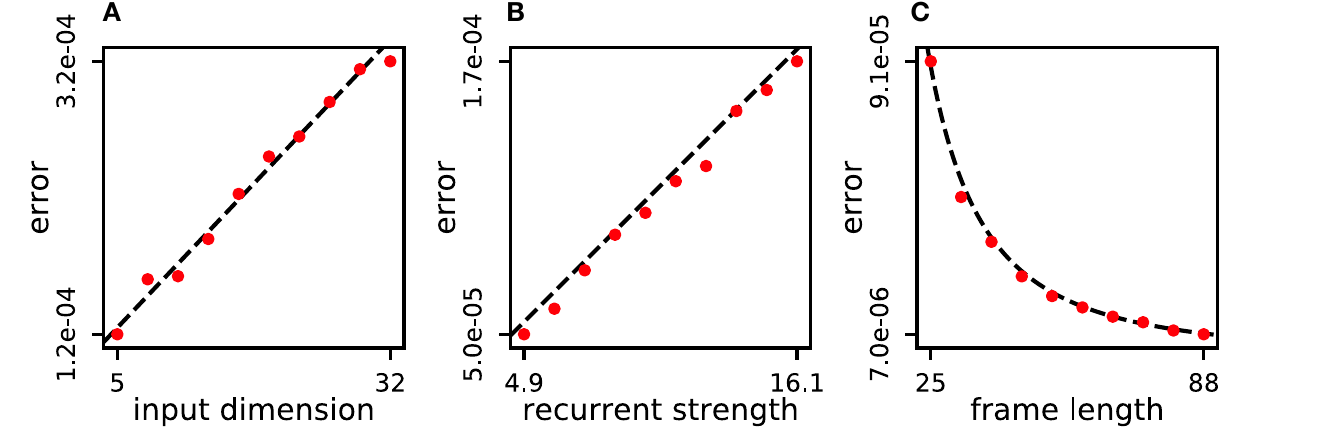}
  \caption{Validating the MSE model by varying parameters of the randomly generated LDS and comparing theory (dashed lines) to numerical experiment (red dots). \textbf{(A)} Varying the input dimension. \textbf{(B)} Varying the recurrent strength of the dynamics matrix. \textbf{(C)} Varying the frame length.}
  \label{fig:threeplots}
\end{figure}

 
Next, we independently varied three parameters of the randomly generated LDS and computed the MSE (Eq. \ref{eq:samplecov}) between the spiking and non-spiking LDS state sequences, the theoretical value of which is obtained by taking the trace of both sides of Eq. \ref{eq:errmodel}. We varied the input dimension $n$, the ``recurrent strength'' of $A$, defined as the trace of the matrix term in Eq. \ref{eq:errmodel}, and the frame length $\ell$ (Fig. \ref{fig:threeplots}A-C). In all three cases, we modified the random LDS generation to provide a sequence of similar LDSs and inputs. We again observed good agreement between theory and experiment.

\subsection{Spiking KF \& application to brain-machine interfaces}
\label{subsec:tnkf}

\begin{figure}
  \centering
  \includegraphics[width=5.5in]{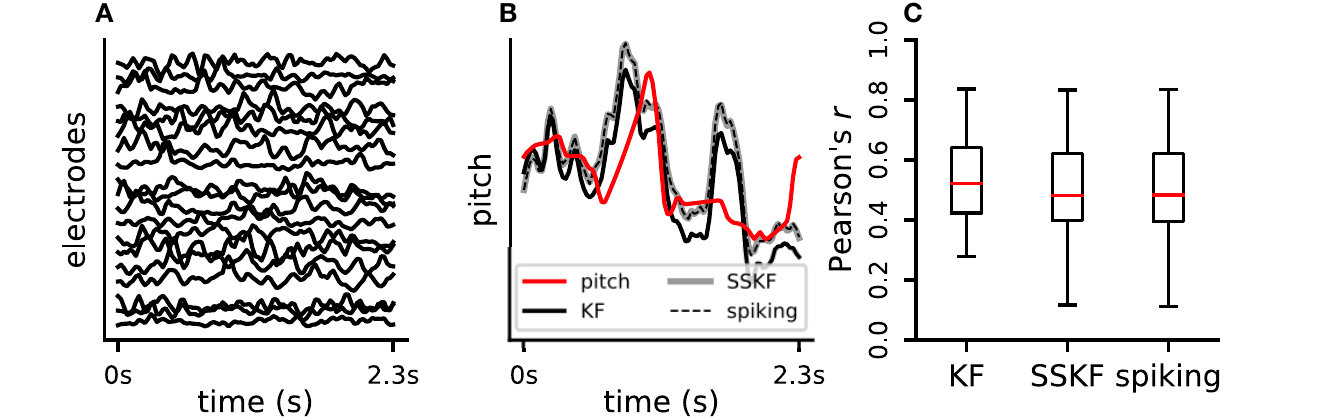}
  \caption{Decoding human vocal pitch from ECoG activity. \textbf{(A)} Selection of ECoG electrodes for a single trial. \textbf{(B)} For the same trial shown in (A), the true pitch alongside reconstructions obtained via the KF, steady-state KF (SSKF), and spiking KF (spiking). \textbf{(C)} Box plot of the Pearson correlations between the true pitch and the three reconstructions across trials. One outlier trial was excluded.}
  \label{fig:tnkf}
\end{figure}

The spiking LDS may be used to run a steady-state KF, which we leveraged to decode human vocal pitch from simultaneously recorded ECoG neural activity (Fig. \ref{fig:tnkf}A). Let $\bo{x}_t \in \mathbb{R}^m$ and $\bo{y}_t \in \mathbb{R}^n$ denote the vocal pitch state and ECoG recording, respectively, at time-step $t$. As per the KF framework, we model the pitch dynamics as linear and Markovian with Gaussian noise, and the ECoG recording as a linear function of the pitch state with Gaussian noise. The Gaussian posterior distribution $P(\bo{x}_t \: | \: \bo{y}_1, \ldots, \bo{y}_t )$ is of interest. If the pitch dynamics are asymptotically stable, then in the limit $t \rightarrow \infty$, the mean $\hat{\bo{x}}_t$ of the posterior evolves according to an asymptotically stable LDS driven by the ECoG data: $\hat{\bo{x}}_t = A_\text{SSKF} \hat{\bo{x}}_{t-1} + B_\text{SSKF} \bo{y}_t$, where $ A_\text{SSKF}$ and $B_\text{SSKF}$ may be computed based on the KF model parameters \cite{chui2017kalman}. The pitch time-series for each trial was augmented with a velocity component as well as a constant bias component. Both the pitch and ECoG data were normalized to $[-\eta p \ell, \eta p \ell]$, where we used $\eta = 0.9, p = 21$, and $\ell = 25$. Note that the 25 ms frame length aligns with the 40 Hz rate of the neural and pitch data, as it would need to in a real-time BMI. For each of the 38 trials, we fit the KF model to the data from all other trials using maximum likelihood techniques, then used a full KF (not in the steady-state limit), a steady-state KF, and a spiking steady-state KF to reconstruct the original pitch trajectory (Fig. \ref{fig:tnkf}B) and computed for each trial the Pearson correlations between the true pitch trajectory and the three reconstructed trajectories (Fig. \ref{fig:tnkf}C).  While the KF-based decoders demonstrated only moderate success at decoding pitch kinematics, presumably due to the limited information about vocal pitch available in the ECoG data, we observed very good alignment between the spiking and non-spiking versions of the steady-state KF. This implies that neuromorphic architectures are a viable low-power deployment platform for portable, KF-based BMIs. See Appendix \ref{sec:appdxkf},\ref{sec:appdxecog} for details.

\section{Discussion}

\par
A neural version of the KF has previously been proposed, but without spiking neurons \cite{NIPS2009_3665}. The Neural Engineering Framework (NEF) also provides a method for mapping LDSs onto spiking neural networks, but assumes rich features of neurons and spikes that are not available on TrueNorth \cite{eliasmith2004neural}. Previous work has suggested that nonlinear dynamical systems may be effective BMI decoders \cite{sussillo2012recurrent}. Whether or not training methods for emulating nonlinear dynamical systems in spiking neural networks (e.g., FORCE learning) may be used for TrueNorth is an important direction for future research \cite{sussillo2009generating,nicola2017supervised,depasquale2018full}.

\par 
Our work provides a complete implementation of a BMI decoder on existing neuromorphic hardware, opening the door to low-power neural prostheses. To reduce power consumption further, we propose pinpointing electrodes which are important for control, allowing unselected electrodes to be turned off. Recent work also suggests that spike sorting is unnecessary for parsing motor cortical dynamics, and thus may safely be skipped to achieve power savings without loss in decoding performance \cite{Trautmann229252}.

\par
Many studies have revealed that motor cortical activity is largely explained by linear dynamics \cite{churchland2012neural, shenoy2013cortical}. Our work reveals, and demonstrates a means of resolving, a key tension between the nonnegative character of spike codes and the mixed-sign nature of general linear dynamics. Our work also demonstrates that the stability of the implemented linear dynamics allows errors due to spiking computation to be forgotten exponentially quickly rather than compounded across time. 

\subsubsection*{Acknowledgments}
We would like to thank Ben Dichter for collecting the ECoG data as well as Rebecca Carney for her contributions to this project.

\bibliography{references}

\input{nips_2018_appendix}

\end{document}

%% file: nips_2018_appendix.tex
\pagebreak

\setcounter{equation}{0}
\setcounter{figure}{0}
\setcounter{table}{0}
\setcounter{section}{0}
\setcounter{page}{1}

\renewcommand{\thesection}{\Alph{section}}
\renewcommand{\thesubsection}{\Alph{section}.\roman{subsection}}
\renewcommand*{\thepage}{A\arabic{page}}
\numberwithin{equation}{section}
\numberwithin{figure}{section}

\section*{Appendix for ``Spiking Linear Dynamical Systems on Neuromorphic Hardware for Low-Power Brain-Machine Interfaces''}

References to equations, figures, and sections which do not begin with a letter point to the main text; otherwise, they point internally to the Appendix.

\section{Statistics of the error due to spiking computation}

\subsection{Scalar multiplication}
\label{subsec:appdxscalarmult}

\par 
Consider a single neuron which follows the model of Eq. \ref{eq:neuron} with a single input. Such a neuron is an unbiased estimator of scalar multiplication by $w = \frac{\alpha}{\beta}$, where $\alpha$ is the ingoing synaptic weight and $\beta$ is the firing threshold. Here, we prove unbiasedness, and derive the autocovariance of the error. Let the variable $t$ index frames (not individual discrete time-steps) and let $V_t$ denote the neuron's membrane potential at the \textit{end} of frame $t$. If $n^\text{in}_t$ is the number of input spikes received by the neuron over frame $t$, then the number $n_t^\text{out}$ of spikes fired by the neuron over the same frame is
\begin{equation}
n^{\text{out}}_t = \left\lfloor \frac{V_{t-1} + \alpha n^\text{in}_t}{\beta} \right\rfloor
\label{eq:output}
\end{equation}
and $V_t$ is propagated according to
\begin{equation}
V_t = V_{t-1} + \alpha n^\text{in}_t - \beta n^{\text{out}}_t.
\label{eq:propagate}
\end{equation}
One subtlety is that, in actuality, $n^\text{out}_t$ in Eq. \ref{eq:output} gives the number of times the neuron will fire \textit{eventually} if provided with $n^\text{in}_t$ spikes -- in general, we are not guaranteed that all $n^\text{out}_t$ spikes will be fired before the end of frame $t$. For this to be guaranteed independent of the timing of the input spikes, we need $w \leq 1$. In this work, we assume that all $n^\text{out}_t$ spikes are fired before the end of frame $t$, regardless of the value of $w$. Our numerical experiments used matrices with all elements less than 1, so Eq. \ref{eq:output} was adhered to exactly.

\par 
The error due to spiking computation at frame $t$ is $\epsilon_t = n^\text{out}_t - wn^\text{in}_t$. We will treat the input sequence $n^\text{in}_t$ as random and make an assumption about its distribution. First, define $r : \mathbb{R} \rightarrow [0, 1)$ as
\begin{equation}
r(x) = x - \left\lfloor x \right\rfloor,
\label{eq:remainder}
\end{equation}
which picks out the non-integer remainder of its argument $x$. Then, we assume that the input sequence satisfies
\begin{equation}
r\left(wn_t^\text{in}\right) \overset{\text{i.i.d.}}{\sim} \mathcal{U}(0, 1),
\label{eq:ass1}
\end{equation}
i.e. that the non-integer remainders of the error-free multiplication results are uniformly distributed on the unit interval and mutually independent across time. Note that Eq. \ref{eq:ass1} does \textit{not} require the input to lack structure on the scale of its full dynamic range $[-\ell, \ell]$. Rather, it requires the input to lack structure on the scale of unity, which is a weak assumption for $\ell \gg 1$. Define the variable $R_t$ as
\begin{equation}
R_t = r\left(w\sum_{i=1}^t n^\text{in}_i \right).
\label{eq:bigrdef}
\end{equation}
Note that the $R_t$'s satisfy the recurrence relation
\begin{equation}
R_t = r\left( R_{t-1} + wn_t^\text{in}\right).
\label{eq:recurrel}
\end{equation}
The main statistical assumption Eq. \ref{eq:ass1} implies that
\begin{equation}
R_t \overset{\text{i.i.d.}}{\sim} \mathcal{U}(0, 1).
\label{eq:bigrass}
\end{equation}
We may show via induction that $\frac{V_t}{\beta} = R_t$. This is easy to check in the base case of $t= 1$. The inductive step for general $t$ is then
\begin{equation}
\begin{split}
\frac{V_t}{\beta} &= \frac{1}{\beta}\left( V_{t-1} + \alpha n^\text{in}_t - \beta \left\lfloor \frac{V_{t-1} + \alpha n^\text{in}_t}{\beta} \right\rfloor \right)\\
&= \frac{V_{t-1}}{\beta} + wn^\text{in}_t - \left\lfloor \frac{V_{t-1} + \alpha n^\text{in}_t}{\beta} \right\rfloor\\
&= R_{t-1} + w n_t^\text{in} - \left\lfloor R_{t-1} + wn_t^\text{in}\right\rfloor \\
&= r\left( R_{t-1} + w n_t^\text{in} \right)\\
&= R_t.
\label{eq:indstep}
\end{split}
\end{equation}
Thus we may write $\epsilon_t$ as
\begin{equation}
\begin{split}
\epsilon_t &= \left\lfloor \frac{V_{t-1} + \alpha n_t^\text{in}}{\beta} \right\rfloor  - wn^\text{in}_t \\
&= \left\lfloor R_{t-1} + wn^\text{in}_t \right\rfloor  - wn^\text{in}_t \\
&= R_{t-1} + wn^\text{in}_t - r\left(R_{t-1} + wn^\text{in}_t\right) - wn_t^\text{in}\\
&= R_{t-1} - r\left(R_{t-1} + wn^\text{in}_t\right) \\
&= R_{t-1} - R_t.
\end{split}
\label{eq:epsilonrewrite}
\end{equation}
Equipped with Eq. \ref{eq:bigrass} and Eq. \ref{eq:epsilonrewrite}, we may calculate the statistics of $\epsilon_t$ up to any order. In particular, the mean is
\begin{equation}
\mathbf{E}\left[ \epsilon_t \right] = \bo{E}\left[R_{t-1}\right] - \bo{E}\left[R_{t}\right] = 0
\label{eq:epsmean}
\end{equation}
and the autocovariance is
\begin{equation}
\text{Cov}\left( \epsilon_{t+\Delta t}, \epsilon_{t}\right) = \bo{E}\left[R_{t + \Delta t - 1}R_{t-1} \right] - \bo{E}\left[R_{t + \Delta t - 1}R_t \right] - \bo{E}\left[R_{t + \Delta t }R_{t-1} \right] + \bo{E}\left[R_{t + \Delta t }R_t \right].
\label{eq:secorder}
\end{equation}
Each of these terms (in absolute value) is either $\frac{1}{4}$ (if the $R_t$'s are distinct) or $\frac{1}{3}$ (if the $R_t$'s are the same). For $\Delta t = 0$, Eq. \ref{eq:secorder} becomes
\begin{equation}
\begin{split}
\text{Cov}\left( \epsilon_{t}, \epsilon_{t}\right) &= \bo{E}\left[R_{t-1}^2 \right] - \bo{E}\left[R_{t-1}R_t \right] - \bo{E}\left[R_{t}R_{t-1} \right] + \bo{E}\left[R_{t}^2 \right]\\
&= \frac{1}{3} - \frac{1}{4} - \frac{1}{4} + \frac{1}{3}\\
&= \frac{1}{6}.
\end{split}
\end{equation}
For $\Delta t = 1$, Eq. \ref{eq:secorder} becomes
\begin{equation}
\begin{split}
\text{Cov}\left( \epsilon_{t+1}, \epsilon_{t}\right) &= \bo{E}\left[R_{t}R_{t-1} \right] - \bo{E}\left[R_{t}^2 \right] - \bo{E}\left[R_{t+1}R_{t-1} \right] + \bo{E}\left[R_{t+1}R_t \right]\\
&= \frac{1}{4} - \frac{1}{3} - \frac{1}{4} + \frac{1}{4}\\
&= -\frac{1}{12}
\end{split}
\end{equation}
Finally, for $\Delta t = 2$, Eq. \ref{eq:secorder} becomes
\begin{equation}
\begin{split}
\text{Cov}\left( \epsilon_{t+2}, \epsilon_{t}\right) &= \bo{E}\left[R_{t+1}R_{t-1} \right] - \bo{E}\left[R_{t+1}R_t \right] - \bo{E}\left[R_{t+2}R_{t-1} \right] + \bo{E}\left[R_{t+2}R_t \right]\\
&= \frac{1}{4} - \frac{1}{4} - \frac{1}{4} + \frac{1}{4}\\
&= 0
\end{split}
\end{equation}
and the autocovariance also vanishes for all $ \Delta t \geq 2$. In summary,
\begin{equation}
\bo{E}\left[ \epsilon_t \right] = 0,\:\:\: \text{Cov}\left(\epsilon_{t+\Delta t}, \epsilon_t\right) = \frac{n}{6}f(\Delta t),
\label{eq:scalarerr}
\end{equation}
where $f(\Delta t)$ is given by Eq. \ref{eq:f}. Using the same strategy, we could calculate higher-order statistics of $\epsilon_t$. For example, $\bo{E}\left[\epsilon_t \epsilon_{t+1} \epsilon_{t+2} \right] = 0$.

\subsection{Matrix-vector multiplication}
\label{subsec:appdxmatmult}

\begin{figure}
  \centering
  \includegraphics[width=5.5in]{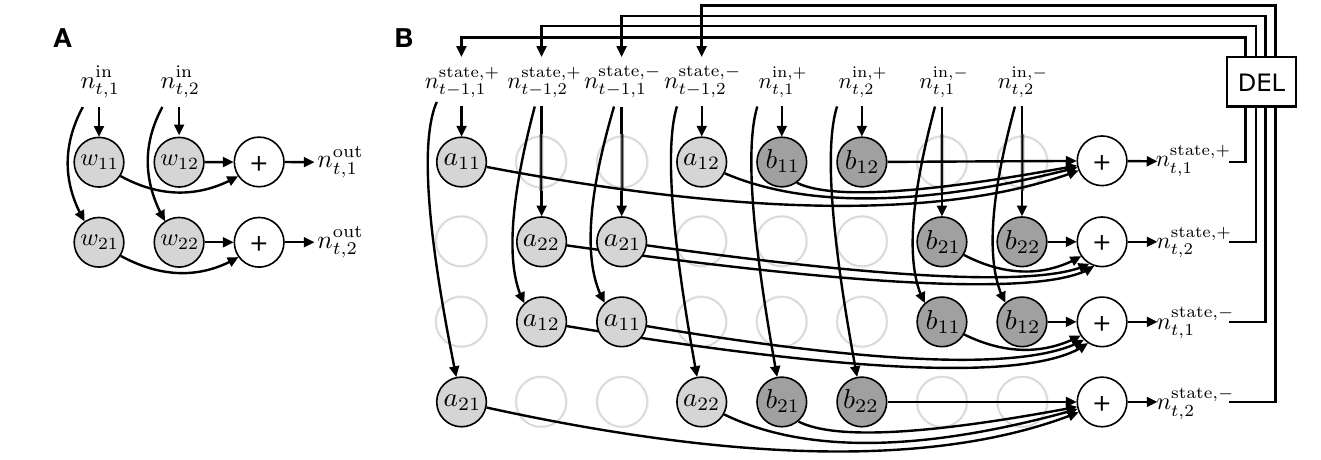}
  \caption{Neural circuit diagrams for linear computations. \textbf{(A)} Nonnegative matrix-vector multiplication for a 2-by-2 matrix. \textbf{(B)} Linear dynamical system with a 2-dimensional state and 2-dimensional input. The dynamics matrix $A$ satisfies $\text{sgn}(A) = ((+, -),(-, +))$ and the input matrix $B$ satisfies $\text{sgn}(B) = ((+, +),(-, -))$.}
  \label{fig:linearcircuits}
\end{figure}

The main statistical assumption of the scalar multiplication case (Eq. \ref{eq:ass1}) must be generalized to obtain the error statistics of the matrix-vector multiplication circuit stated in Eq. \ref{eq:vecstats}-\ref{eq:f} (Fig. \ref{fig:linearcircuits}A shows a diagram of this circuit). In particular, we assume that, for all $t,i,j$,
\begin{equation}
r\left( w_{ij}n^\text{in}_{t,j} \right)  \overset{\text{i.i.d.}}{\sim} \mathcal{U}(0, 1).
\label{eq:finalass}
\end{equation}

\subsection{Linear dynamical systems}
\label{subsec:appdxlds}

The statistical assumption for matrix-vector multiplication (Eq. \ref{eq:finalass}) must be generalized yet again to obtain the error statistics of the LDS circuit stated in Eq. \ref{eq:statsmixedldsmaintext} (Fig. \ref{fig:linearcircuits}B shows a diagram of this circuit). Let $\tilde{a}_{ij}$ and $\tilde{b}_{ij}$ denote the elements of the transformed system matrices $\tilde{A}$ and $\tilde{B}$, respectively, and let $\bo{n}^\text{in}_t = \left(\begin{array}{cc} \bo{n}^{\text{in},+}_t & \bo{n}^{\text{in},-}_t \end{array}\right)$ and $\bo{n}^\text{state}_t = \left(\begin{array}{cc} \bo{n}^{\text{state},+}_t & \bo{n}^{\text{state},-}_t \end{array}\right)$. We assume that, for all $t, t', i, j, k, \ell$ such that that $\tilde{a}_{ij} \neq 0, \tilde{b}_{ij} \neq 0$, and $n^\text{in}_{t', \ell} \neq 0$,
\begin{subequations}
\begin{align}
r(\tilde{a}_{ij}n^\text{state}_{t,j}) &\overset{\text{i.i.d.}}{\sim} \mathcal{U}(0, 1) \\
r(\tilde{b}_{k \ell}n^\text{in}_{t',\ell}) &\overset{\text{i.i.d.}}{\sim} \mathcal{U}(0, 1) 
\end{align}
\end{subequations}
where mutual independence is required amongst the whole set of $r(\tilde{a}_{ij}n^\text{state}_{t,j})$ and $r(\tilde{b}_{k \ell}n^\text{in}_{t',\ell})$ variables.

\section{Spiking linear dynamical systems}

\subsection{Proof of spiking system matrix identities }
\label{subsec:appdxidproof}

Let $\bo{v} \in \mathbb{R}^{2m}$ and $k \geq 1$. The identity involving the transformed dynamics matrix $\tilde{A}$ (Eq. \ref{eq:identities}) may be proved via induction. In the base case of $k = 1$, we have
\begin{equation}
\tilde{A}\bo{v} = \tbomat{\text{ReLU}(+A)\left[\bo{v}\right]^+ - \text{ReLU}(-A)\left[\bo{v}\right]^- }{ \text{ReLU}(-A)\left[\bo{v}\right]^+ - \text{ReLU}(+A)\left[\bo{v}\right]^- }
\end{equation}
and so
\begin{equation}
\begin{split}
\left[\tilde{A}\bo{v} \right]^+ - \left[\tilde{A}\bo{v} \right]^- &= 
\text{ReLU}(A)\left[\bo{v}\right]^+ - \text{ReLU}(-A)\left[\bo{v}\right]^- - \text{ReLU}(-A)\left[\bo{v}\right]^+ + \text{ReLU}(A)\left[\bo{v}\right]^- \\
&= \left(\text{ReLU}(A) - \text{ReLU}(-A) \right) \left(\left[\bo{v}\right]^+ - \left[\bo{v}\right]^- \right)\\
&= A\left(\left[\bo{v}\right]^+ - \left[\bo{v}\right]^- \right).
\end{split}
\end{equation}
The inductive step is then
\begin{equation}
\begin{split}
\left[\tilde{A}^{k+1}\bo{v} \right]^+ - \left[\tilde{A}^{k+1}\bo{v} \right]^-
&= A\left(\left[\tilde{A}^k\bo{v}\right]^+ - \left[\tilde{A}^k\bo{v}\right]^- \right)\\
&= A^{k+1}\left(\left[\bo{v}\right]^+ - \left[\bo{v}\right]^- \right).
\end{split}
\end{equation}
The proof of the identity involving $\tilde{B}$ does not require induction and is similar to the $k=1$ case for $\tilde{A}$.

\subsection{Perfect recovery of original LDS state}
Using Eq. \ref{eq:biglds}, the state of the spiking LDS at frame $t$ is 
\begin{equation}
\tbomat{\bo{n}^{\text{state},+}_t}{\bo{n}^{\text{state},-}_t} = \sum_{k=0}^{t-1}\tilde{A}^k\tilde{B} \tbomat{\bo{n}^{\text{in},+}_{t-k}}{\bo{n}^{\text{in},-}_{t-k}} + \sum_{k=0}^{t-1}\tilde{A}^k \tbomat{\bs{\epsilon}^{+}_{t-k}}{\bs{\epsilon}^{-}_{t-k}}.
\end{equation}
If the error terms are zero, then
\begin{equation}
\begin{split}
\bo{n}^{\text{state},+}_t - \bo{n}^{\text{state},-}_t
&= \left[\sum_{k=0}^{t-1}\tilde{A}^k\tilde{B} \tbomat{\bo{n}^{\text{in},+}_{t-k}}{\bo{n}^{\text{in},-}_{t-k}}\right]^+ - \left[\sum_{k=0}^{t-1}\tilde{A}^k\tilde{B} \tbomat{\bo{n}^{\text{in},+}_{t-k}}{\bo{n}^{\text{in},-}_{t-k}}\right]^-\\
&= \sum_{k=0}^{t-1} \left(\left[ \tilde{A}^k\tilde{B} \tbomat{\bo{n}^{\text{in},+}_{t-k}}{\bo{n}^{\text{in},-}_{t-k}} \right]^+ - \left[ \tilde{A}^k\tilde{B} \tbomat{\bo{n}^{\text{in},+}_{t-k}}{\bo{n}^{\text{in},-}_{t-k}} \right]^- \right)\\
&= \sum_{k=0}^{t-1}A^k B \left( \bo{n}^{\text{in},+}_{t-k} - \bo{n}^{\text{in},-}_{t-k} \right) \\
&= \sum_{k=0}^{t-1}A^k B \left( \text{ReLU}(\bo{u}_{t-k}) - \text{ReLU}(-\bo{u}_{t-k})\right) \\
&= \sum_{k=0}^{t-1}A^k B \bo{u}_{t-k} \\
&= \bo{x}_t.
\label{eq:derivv}
\end{split}
\end{equation}

\subsection{Statistics of the residual}
As stated in the main text, the residual at frame $t$ between the spiking and non-spiking state sequences is
\begin{equation}
\bo{r}_t = \left[ \bo{n}^{\text{state},+}_t - \bo{n}^{\text{state},-}_t \right] - \bo{x}_t = \sum_{k=0}^{t-1}A^k\Delta \bs{\epsilon}_{t-k},
\label{eq:resid}
\end{equation}
where $\Delta \bs{\epsilon}_{t} = \bs{\epsilon}^+_{t} - \bs{\epsilon}^-_{t}$. This may be proved using a derivation similar to Eq. \ref{eq:derivv}, but without neglecting the error terms $\bs{\epsilon}^+_t$ and $\bs{\epsilon}^+_t$ . The first- and second-order statistics of $\Delta \bs{\epsilon}_{t}$ are given by Eq. \ref{eq:statsmixedldsmaintext}. Since $\bo{E}\left[\Delta \bs{\epsilon}_{t}\right] = 0$,  it follows from Eq. \ref{eq:resid} that $\bo{E}\left[\bo{r}_t\right] = 0$. The covariance of $\bo{r}_t$ is
\begin{equation}
\text{Cov}(\bo{r}_t) = \sum_{j=0}^{t-1}\sum_{k=0}^{t-1}A^j \text{Cov}\left(\Delta \bs{\epsilon}_{t-j}, \Delta\bs{\epsilon}_{t-k}\right)\left(A^k\right)^T.
\label{eq:covvv}
\end{equation}
Now, $\text{Cov}\left(\Delta \bs{\epsilon}_{t-j}, \Delta\bs{\epsilon}_{t-k}\right)$ is nonzero when $\left|j-k\right| \in \{0, 1\}$ according to Eq. \ref{eq:statsmixedldsmaintext}. Substituting the covariance expression of Eq. \ref{eq:statsmixedldsmaintext} into Eq. \ref{eq:covvv} gives
\begin{equation}
\begin{split}
\text{Cov}(\bo{r}_t) &= \frac{2m + n}{6}\sum_{k=0}^{t-1}A^k \left(A^k\right)^T
- \frac{2m + n}{12}A\sum_{k=0}^{t-2}A^k \left(A^k\right)^T
- \frac{2m + n}{12}\sum_{k=0}^{t-2}A^k \left(A^k\right)^T A^T\\
&= \frac{2m + n}{6}\text{sym}\left(\sum_{k=0}^{t-1}A^k \left(A^k\right)^T - A\sum_{k=0}^{t-2}A^k \left(A^k\right)^T\right).
\end{split}
\label{eq:coveq}
\end{equation}
As $t \rightarrow \infty$, the different upper limits of the sums in Eq. \ref{eq:coveq} are not relevant and we have
\begin{equation}
\text{Cov}\left( \bo{r}_{t \rightarrow \infty} \right) = \frac{2m + n}{6}\text{sym}\left(\left(I - A\right) \sum_{k=0}^\infty A^k\left(A^k\right)^T \right).
\label{eq:sscov}
\end{equation}
The full autocovariance is, for $\Delta t \geq 0$,
\begin{equation}
\text{Cov}\left( \bo{r}_{t + \Delta t}, \bo{r}_t \right) = \frac{2m + n}{6}A^{\Delta t}\left(\sum_{k=0}^{t-1-\Delta t}A^k \left(A^k\right)^T
- \frac{1}{2} A\sum_{k=0}^{t-2 - \Delta t}A^k \left(A^k\right)^T
- \frac{1}{2}  \sum_{k=0}^{t - \Delta t}A^k \left(A^k\right)^T A^T \right).
\end{equation}
As $t \rightarrow \infty$, the different limits of the sums are again not relevant and we have
\begin{equation}
\text{Cov}\left( \bo{r}_{t + \Delta t}, \bo{r}_t \right) = A^{\Delta t}\text{Cov}\left(\bo{r}_{t} \right) \:\:\: \left(\Delta t \geq 0, \: t \rightarrow \infty \right).
\end{equation}

\subsection{Closed-form expression for matrix series}
The steady-state covariance matrix of the residual between the spiking and non-spiking LDS state sequences (Eq. \ref{eq:mixedcovinf}, reproduced here as Eq. \ref{eq:sscov}) contains a matrix series in $A$. In order to calculate this, it is preferable to use a closed-form expression as opposed to a truncated sum. If $A$ is diagonalizable, then we may write $A = V D V^{-1}$ for some complex matrices $V$ and $D = \text{diag}\left(\bs{\lambda} \right)$, where $\bs{\lambda}$ is a column vector of the eigenvalues. Then,
\begin{equation}
\begin{split}
\sum_{k=0}^{\infty} A^k \left(A^k\right)^T &= \sum_{k=0}^\infty VD^k V^{-1}\left(V^{-1}\right)^T D^k V^T\\
&= V \left( \sum_{k=0}^\infty  D^k \left(V^T V\right)^{-1}D^k \right) V^T\\
&= V \left(  \left(V^T V\right)^{-1} \circ \sum_{k=0}^\infty \left(\bs{\lambda}\bs{\lambda}^T \right)^k \right) V^T\\
&= V \left( \left(V^T V\right)^{-1} \circ \frac{1}{1 - \bs{\lambda}\bs{\lambda}^T} \right) V^T.
\end{split}
\end{equation}
where all operations on the outer product $\bs{\lambda}\bs{\lambda}^T$ are element-wise.

\subsection{Serial correlations of residuals}
\label{subsec:appdxserialcorr}

\par 
The timescale $\tau_A$ provides a rough idea of the minimum power $k$ such that $A^k$ small. In particular, for $0 < \epsilon < 1$, 
\begin{equation}
\rho(A)^k = \epsilon \implies k = \frac{\log 1/ \epsilon}{\log 1/\rho(A)} \implies \tau_A \sim  \frac{1}{\log 1 / \rho(A)} .
\end{equation}
This means that if one runs the spiking LDS and the equivalent non-spiking LDS for $T$ frames and computes the residuals between the two state sequences, one does not obtain $T$ i.i.d. samples of the residuals. Rather, one obtains some smaller effective number $n_\text{eff} < T$ of i.i.d. samples, where $n_\text{eff} \sim \frac{T}{\tau_A}$. We accounted for this when numerically validating the predicted residual covariance by considering time-series lasting $T$ frames where $T \gg \tau_A$.

\subsection{Necessary \& sufficient condition for stability of the spiking system}
\label{subsec:appdxnecsuf}

\par 
The nonnegative, twice-as-large dynamics matrix of the spiking LDS is given by
\begin{equation}
\tilde{A} = \tbtmat{\relu{A}}{\relu{-A}}{\relu{-A}}{\relu{A}}.
\end{equation}
The spiking system is asymptotically stable if and only if $\rho(\tilde{A}) < 1$. Note that for any square matrices $V$ and $W$, there is a determinant identity
\begin{equation}
\left| \begin{array}{cc} V & W \\ W & V \end{array} \right|
= \left|V - W \right| \left|V + W \right|.
\end{equation}
It follows from this identity that the characteristic polynomial of $\tilde{A}$ factorizes into those of $A$ and $\text{abs}(A)$, the element-wise absolute value of $A$:
\begin{equation}
\begin{split}
p_{\tilde{A}}(\lambda) &= \left| \begin{array}{cc}
    \text{ReLU}(A) - \lambda I_m & \text{ReLU}(-A) \\ 
    \text{ReLU}(-A) & \text{ReLU}(A) - \lambda I_m 
 \end{array} \right| \\
 &= \left| \text{ReLU}(A) + \text{ReLU}(-A) -\lambda I_m \right|
 \left| \text{ReLU}(A) - \text{ReLU}(-A) -\lambda I_m \right|\\
 &= \left| \text{abs}(A) - \lambda I_m \right| \left| A - \lambda I_m \right|\\
 &= p_{\text{abs}(A)}\left(\lambda\right) p_{A}\left(\lambda\right)
\end{split}
\end{equation}
Thus, the spectrum of $\tilde{A}$ is the union of those of $A$ and $\text{abs}(A)$, and a necessary and sufficient condition for $\rho(\tilde{A}) < 1$ is $\rho({A}) < 1$ \textit{and} $\rho(\text{abs}({A})) < 1$. However, note that $\rho(\text{abs}({A})) < 1$ implies $\rho({A}) < 1$. To prove this, note that, for any square matrices $V$ and $W$, we have
\begin{equation}
\left|VW\right| \leq \left|V\right|\left|W\right|.
\end{equation}
Setting $V = W = A$ gives
\begin{equation}
\left| A^2 \right| \leq \left| A\right|^2.
\label{eq:ineq}
\end{equation}
We may apply Eq. \ref{eq:ineq} repeatedly to show that $\left| A^k \right| \leq \left| A \right|^k$ for all $k \geq 1$. Assuming $\rho(\text{abs}( A)) < 1$, we have that $\lim_{k \rightarrow \infty} \text{abs}( A)^k = 0$. It follows that $\lim_{k \rightarrow \infty} A^k = 0$, so $\rho\left( A \right) < 1$. Therefore, a necessary and sufficient condition for $\rho(\tilde{A}) < 1$ is simply $\rho(\text{abs}(A)) < 1$.

\section{Mapping onto TrueNorth}
\label{sec:appdxtnmap}

\begin{figure}
  \centering
  \includegraphics[width=2.16in]{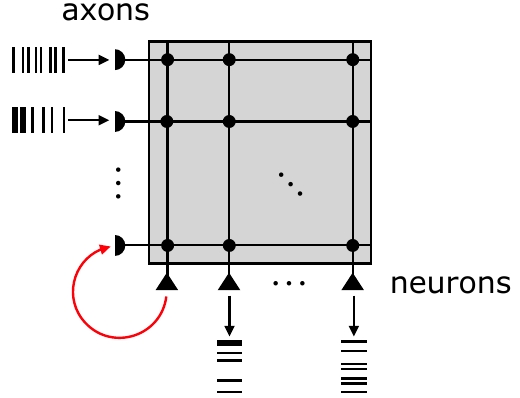}
  \caption{Representation of a single TrueNorth core. 256 presynaptic axons (semicircles connected to horizontal lines) run vertically while 256  neurons (triangles) run horizontally. Each neuron has a single dendrite (vertical line) and each axon is able to synapse with any dendrite on the same core. A synapse, denoted by a dot at the intersection of an axon and a dendrite, is associated with a synaptic weight. Together, the synaptic weights define a synaptic weight matrix. Axons receive spikes, which travel rightward along the horizontal lines. Upon hitting a synapse, spikes are directed downward along a dendrite until they are integrated by a neuron. When a neuron fires, its spikes are routed to exactly one axon on any core, including the neuron's own core (red arrow). TrueNorth contains 4,096 cores joined in a 64-by-64 mesh network.}
  \label{fig:tncore}
\end{figure}

\subsection{\textit{\textbf{p}}-Dimensional Scalar Multiplication}
\label{subsec:tnscalarmult}

\begin{figure}
  \centering
  \includegraphics[width=5.5in]{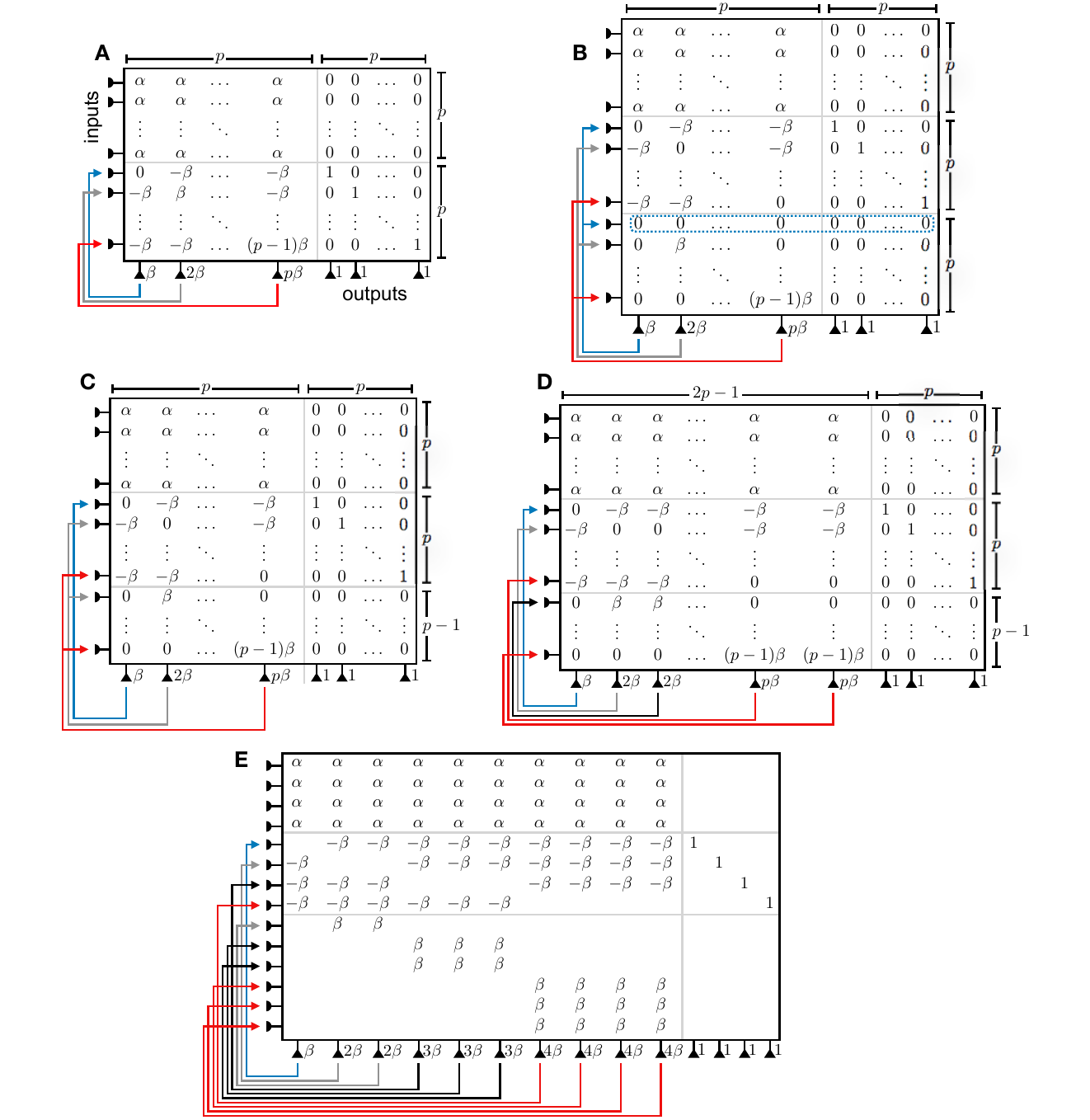}
  \caption{Mapping the $p$-dimensional scalar multiplication circuit onto a TrueNorth core. See Section \ref{subsec:tnscalarmult} for full description. The color gradients of the recurrent connections are for purely visual purposes. \textbf{(A)} The most natural crossbar, which is limited to $p \leq 3$ due to the limited number of axon types. \textbf{(B)} Crossbar from (A), but with the recurrent part of the crossbar separated into its on- and off-diagonal parts. This crossbar is not viable since neurons route spikes to pairs of axons. A functionless row of all zeros is circled in blue. \textbf{(C)} Crossbar from (B) but with the row of all zeros removed. \textbf{(D)} Crossbar from (C), but with neurons duplicated to route spikes to pairs of axons. The parameter $\beta$ is limited for large $p$ since we must instantiate the synaptic weight $(p-1)\beta$, and synaptic weights may be at most 255. \textbf{(E)} Crossbar from (D), but with neuron and axon replications which allow for $\beta$ to be as large as 255 for $p \leq 21$. }
  \label{fig:mult_crossbars}
\end{figure}

\par 
In order to implement the $p$-dimensional multiplication circuit described in the Section 3.5 of the main text, we must translate the circuit of Fig. 1B into a TrueNorth synaptic crossbar (Fig. \ref{fig:tncore}). A similar problem, with essentially the same solution, arises when implementing the $p$-dimensional versions of the addition and cancellation circuits on TrueNorth. This crossbar, shown in Fig. \ref{fig:mult_crossbars}E, is somewhat complicated and makes little sense on its own. Rather, it is best understood as the result of a series of workarounds to TrueNorth's constraints, described in the main text. In Fig. \ref{fig:mult_crossbars}, the progression from a scalar multiplication crossbar which violates TrueNorth's constraints to one that is compatile with TrueNorth is shown. Here, we walk through this series of crossbars step-by-step.

\par 
We start with the crossbar of Fig. \ref{fig:mult_crossbars}A. The $p$-by-$p$ sub-matrix of $\alpha$'s in the upper-left quadrant directs input spikes to the first $p$ neurons, implementing the feedforward, fully-connected part of circuit of Fig. 1B. When these neurons fire, their spikes are routed to the second set of $p$ axons, which in turn direct the spikes recurrently to the first $p$ neurons via the lower-left quadrant of the crossbar. This implements the recurrent connections of Fig. 1B. Finally, the identity matrix in the bottom-right quadrant of the crossbar directs these spikes to the second set of $p$ neurons. This means that the second set of $p$ neurons reproduces the output of the first $p$ neurons with a one-time-step delay. The second set of $p$ neurons is not included in the circuit of Fig. 1B, however we include these neurons in the TrueNorth implementation so that the output spikes of the circuit may be routed to axons on any core. Note that these extra neurons increase the latency of the circuit by one time-step.

\par 
This crossbar is incompatible with the parameterization of synaptic weights on TrueNorth. To highlight the issue, let $W^\text{rec}$ be the $p$-by-$p$ sub-matrix in the lower-left quadrant of the crossbar in Fig. \ref{fig:mult_crossbars}A, given by
\begin{equation}
w^\text{rec}_{ij} = \begin{cases}
 (i-1)\beta & i = j \\
 -\beta & i \neq j.
 \end{cases}
 \label{eq:wrec}
\end{equation}
To parameterize $W^\text{rec}$ in the form required by TrueNorth, we must assign an axon type $G_i \in \{0, 1, 2, 3\}$ to each axon $1 \leq i \leq p$. Due to the structure of $W^\text{rec}$, the axon types must satisfy the constraints
\begin{equation}
\begin{split}
G_1, \ldots, G_{i-1}, G_{i+1}, \ldots, G_p \neq G_i, \text{   $1 \leq i \leq p$}.
\end{split}
\label{eq:constraints}
\end{equation}
If the $i$-th constraint is not satisfied, then there is some $j \neq i$ such that $w^\text{rec}_{ij} = w^\text{rec}_{jj}$, contradicting the form of $W^\text{rec}$ (Eq. \ref{eq:wrec}). These constraints are equivalent to the constraint that all the $G_i$'s are distinct. But since their domain contains only four values, we must have $p \leq 4$. Moreover, to implement the whole crossbar of Fig. \ref{fig:mult_crossbars}A, one axon type value is claimed by the sub-matrix of $\alpha$'s, limiting us further to $p \leq 3$, which is unacceptable.

\par
To solve the axon type problem, we decompose $W^\text{rec}$ as the sum of two matrices: the first contains the off-diagonals of $W^\text{rec}$ with zero on-diagonals, while the second contains the on-diagonals of $W^\text{rec}$  with zero off-diagonals. These matrices are then stacked vertically, the second below the first, and substituted for $W^\text{rec}$ in the $p$-dimensional multiplication crossbar (Fig. \ref{fig:mult_crossbars}B). If the $i$-th neuron in this crossbar is allowed to route its spikes to \textit{two} axons, corresponding to the $i$-th rows of both matrices, then the behavior of the original crossbar is preserved. But now, independent of $p$, we need only three distinct axon type values: one for the sub-matrix of $\alpha$'s, one for the sub-matrix containing the off-diagonals of $W^\text{rec}$, and one for the sub-matrix containing the on-diagonals of $W^\text{rec}.$ Note that this crossbar contains a row of all-zeros which serves no purpose. This row is circled by a blue dotted line in Fig. \ref{fig:mult_crossbars}B. The only change in Fig. \ref{fig:mult_crossbars}C is to eliminate this row. 

\par
This crossbar is not viable as it stands since we have violated the requirement that neurons route spikes to exactly one axon. As a workaround, we duplicate columns of the crossbar (Fig. \ref{fig:mult_crossbars}D). Specifically, the column corresponding to each neuron which is supposed to route spikes to two axons is duplicated. A neuron which results from such a column duplication is a functional replica of the original. The ``identical twin" neurons are then capable of routing spikes to the two desired axons. 

\par
With the crossbar of Fig. \ref{fig:mult_crossbars}D, we may implement the $p$-dimensional multiplication circuit on TrueNorth for any $p \leq 85$, where the upper-bound is due to finite core size. However, a serious limitation arises due to the fact that synaptic weights are in $\{-255, \ldots, 255\}$. In the crossbar of Fig. \ref{fig:mult_crossbars}D, the largest weight is $(p-1)\beta$, forcing $\beta \leq \left\lfloor \frac{255}{p-1} \right\rfloor$. This severely limits the number of distinct multipliers, given by $\frac{\alpha}{\beta}$, which may be implemented. Our final modification to the crossbar will allow us to implement values of $\beta$ up to 255 for $p \leq 21$. In the crossbar of \ref{fig:mult_crossbars}D, there are rows in the bottom third of the crossbar which are zero except for two entries equal to $i\beta$ for some $i$. These are the problematic rows which impose the small upper-bound on $\beta$. For $2  \leq i \leq p-1$, we replace the crossbar row containing the weight $i \beta$ with $i$ copies of the row, each with the weight replacement $i\beta \rightarrow \beta$. This creates $i-1$ new axons. If we can supply the new axons with the same spikes supplied to the original axon, then the two relevant neurons will receive effective weight $i\beta$. To supply these spikes, we create $i-1$ new neurons by duplicating columns. In Fig. \ref{fig:mult_crossbars}E, we write down the crossbar for $p = 4$.

\par
One final and easy-to-implement optimization concerns the case of small multipliers. In particular, if $w = \frac{\alpha}{\beta} \leq \frac{1}{p}$, then only the first neuron in the $p$-dimensional multiplication crossbar ever fires. Thus, we may replace the complicated large, complicated crossbar described above with a small, simple one containing one neuron with firing threshold $\beta$ and $p$ axons, each of which synapses with the neuron's dendrite with synaptic weight $\alpha$. In this case, $\beta$ may be as large as the maximum allowable firing threshold on TrueNorth. The register for this parameter has 18 bits, so $\beta$ may be as large as $2^{18}-1 = 282,143$. This optimization is highly beneficial since it both reduces neuronal footprint and increases the accuracy of matrix element representation. To reap the full benefits of the reduced neuronal footprint of multiplication circuits with the small-$w$ optimization, it is important to place many such circuits on single cores.

\subsection{Matrix element representation}
\par
In deriving the spiking LDS, we assumed that the elements of the original LDS system matrices were rational. We then expressed each matrix element (in absolute value) as a ratio $\frac{\alpha}{\beta}$ for integers $\alpha$ and $\beta$ and implemented a multiplication circuit using parameters $\alpha$ and $\beta$. However, the TrueNorth implementation enforces the constraint $\alpha, \beta \in \{0, \ldots, 255\}$. In practice, for a given matrix element $w \geq 0$, we choose $\alpha$ and $\beta$ according to
\begin{equation}
(\alpha, \beta) = \argmin_{ (\alpha', \beta') \in \{0, \ldots, \alpha_\text{max}\} \times \{1, \ldots, \beta_\text{max}\} } \left( w - \frac{\alpha'}{\beta'} \right)^2
\label{eq:minim}
\end{equation}
where
\begin{equation}
\begin{split}
\alpha_\text{max} &= 255 \\
\beta_\text{max} &= \begin{cases}
 255 & w > \frac{1}{p} \\
 282,143 & w \leq \frac{1}{p} \:\:\: \text{(small-$w$ case, see Section \ref{subsec:tnscalarmult})}.
 \end{cases}
 \end{split}
\end{equation}
To implement a spiking LDS with input dimension $n$ and state dimension $m$, we must solve this optimization problem $nm + m^2$ times, and an efficient algorithm is therefore of interest. When either $\alpha'$ or $\beta'$ is fixed and the free variable is treated as a a continuous quantity, the objective function has a unique global minimum in the free variable which may be computed in constant time. If $x^*$ is the value of the free variable at the minimum, then the integral value of the free variable which minimizes the objective function is either $\left\lfloor x^* \right\rfloor$ or $\left\lceil x^* \right\rceil$. By testing both options, we can find the right one. To perform joint optimization in both variables, we fix one of the variables at each integral value in its domain and, for each value, optimize over the other variable using the aforementioned method. The time complexity of this algorithm is $\mathcal{O}\left(\text{min}\left(\alpha_\text{max}, \beta_\text{max}\right)\right)$. This suits our purposes well, since the runtime becomes independent of $\beta_\text{max}$ when $\beta_\text{max}$ is large in the small-$w$ case. This is a significant improvement over the naive algorithm which tests every solution and has time complexity $\mathcal{O}(\alpha_\text{max} \beta_\text{max})$.

\par 
The parameters $\alpha_\text{max}$ and $\beta_\text{max}$ determine how closely the rational matrix elements used in the spiking LDS match those of the original LDS. It is straightforward to obtain a back-of-the-envelope estimate of how large we expect the difference between a matrix element and its rational approximation to be. For simplicity, let $\alpha_\text{max} = \beta_\text{max} = n$ and assume that $0 \leq w \leq 1$. Define $\mathcal{X} = \left\{\left. \frac{\alpha'}{\beta'} \: \right| \: (\alpha, \beta) \in \{0, \ldots, n\} \times \{1, \ldots, n\},\: \frac{\alpha'}{\beta'} \leq 1 \right\}$. We may approximate the size of this set to be $\left| \mathcal{X} \right| \approx \frac{n^2}{2}$. If we assume that the values in $\mathcal{X}$ are evenly spaced over the unit interval, then the rational approximation of $w$ is equivalent to quantizing $w$ with uniform step size $\Delta = \frac{1}{\left| \mathcal{X} \right|} = \frac{2}{n^2}$. If $w$ is a uniform random variable over the unit interval, then the mean-squared quantization error is $\frac{\Delta^2}{12} = \frac{1}{3n^4}$. We therefore expect a typical error due to the rational approximation to be $\sigma \approx \frac{1}{\sqrt{3}}\frac{1}{n^2}$. Substituting $n = 255$, we obtain $\sigma \sim 10^{-5}$. Thus, we expect the spiking LDS system matrices to be very slightly perturbed versions of those of the original LDS.

\par 
When the spiking LDS system matrices are perturbed versions of those of the original LDS, the spiking LDS provides an unbiased estimate of the state sequence of a non-spiking LDS \textit{with the perturbed matrices}. It is easy to check that small perturbations to the dynamics matrix of an asymptotically stable LDS result in small perturbations of the resulting state sequence. For example, consider a LDS given by $\bo{x}_t = A\bo{x}_{t-1} + B\bo{u}_{t}$ with $\rho(A) < 1$ and a perturbed LDS given by $\bo{x}'_t = A'\bo{x}'_{t-1} + B\bo{u}_t$ where $A' = A + P$ for some matrix $P$. Then, to first order in $P$, 
\begin{equation}
\bo{x}'_t - \bo{x}_t = \sum_{k=0}^{t-1} k A^k P B\bo{u}_{t-k}.
\label{eq:perturb}
\end{equation}
If $P$ has elements of order $\sigma \ll 1$, then the difference $\bo{x}'_t - \bo{x}_t$ is very small compared to the scale of $\bo{x}_t$. For our numerical experiments, the perturbation to the state sequences due to rational weight approximation was negligible compared to the error due to spiking computation.

\subsection{Tree-structured addition \& cancellation circuits}
\par
In the spiking LDS for $p$-dimensional spike trains, we must instantiate addition circuits which combine $N = m+n$ spike trains, each of dimension $p$. However, when either $N$ or $p$ is large, such an addition circuit does not fit on a single TrueNorth core since it requires more than 256 axons. If $k$ is the maximum number of $p$-dimensional spike trains which may be added on a single core, then we must combine several $k$-way addition circuits to form an $N$-way addition circuit. Although there are many ways to do this, we seek, in the interest of minimizing neuronal footprint, the way which minimizes the number of $k$-way adders which must be instantiated. Here, we provide an algorithm for finding the correct configuration for a given $N$ and $k$. Note that essentially the same problem arises when cancellation circuits are used in place of addition circuits. For the sake of simplicity, we restrict here to the addition case. The solution for the cancellation case then follows straightforwardly.

\par
\begin{figure}
  \centering
  \includegraphics[width=5.5in]{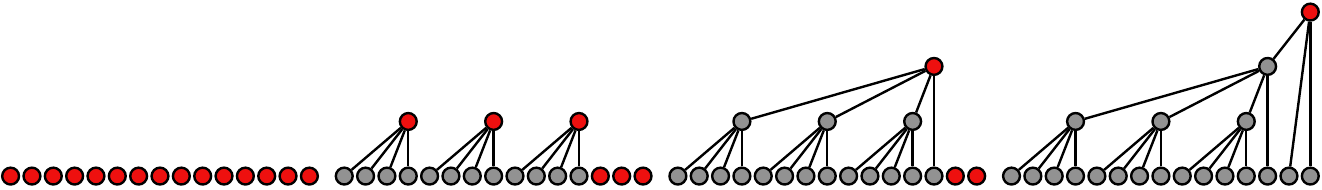}
  \caption{Algorithm for finding a $k$-ary tree, with $N$ leaves, which minimizes the number of internal nodes. Start with a set $\mathcal{S}$ containing the $N$ leaf nodes (nodes in $\mathcal{S}$ are colored red). Then, using nodes in $\mathcal{S}$, form as many groups of exactly $k$ nodes as possible. For each of these groups, make the nodes in the group the children of a new parent. Remove the children from $\mathcal{S}$ and add the parent. Repeat until $|\mathcal{S}| = 1$. For $N = 15$ and $k = 4$, shown here, the algorithm requires four steps. The final tree contains $\left\lceil\frac{N-1}{k-1}\right\rceil = \left\lceil\frac{15 - 1}{4 - 1}\right\rceil = 5$ internal nodes.}
  \label{fig:algo}
\end{figure}

 
The problem outlined above may be stated abstractly as follows: Find a $k$-ary tree which has $N$ leaves and minimizes the number of internal nodes. In this abstract description, the $N$ leaves of the tree correspond to the $N$ input spike trains, the internal nodes correspond to $k$-way adders, and the root corresponds to the final $k$-way adder which outputs the sum of all $N$ inputs. A simple greedy algorithm is optimal. Let $\mathcal{S}$ denote a set of nodes. Initially, $\mathcal{S}$ holds the $N$ leaf nodes. Then, iterate the following procedure: using nodes from $\mathcal{S}$, form as many groups of exactly $k$ nodes as possible. For each group, create a parent node and make the nodes in the group children of the parent. Then, remove the children from $\mathcal{S}$ and add the parent to $\mathcal{S}$. Repeat this process until $|\mathcal{S}| = 1$. This algorithm is illustrated in Fig. \ref{fig:algo}. When the algorithm terminates, the tree contains $\left\lceil \frac{N-1}{k-1} \right\rceil$ internal nodes. Any smaller number of internal nodes cannot form a $k$-ary tree with $N$ leaves.

\par 
Note that in a tree produced by the algorithm, different leaves have difference distances from the root. When we use this algorithm to construct an $N$-way addition circuit, this means the time it takes for a spike to propagate from one of the $N$-way adder inputs to the final output will vary across the inputs. This is undesirable behavior. To account for this, we use a feature on TrueNorth which allows one to configure a neuron such that its spikes are delayed by some number of time-steps before being registered by the receiving axon. By appropriately adjusting the delays of the neurons in the circuit, we can configure the circuit so that the propagation time from leaf to root is the same across leaves.

\section{Validation of error model}
\label{sec:appdxldsgen}

\subsection{$\bo{\eta}$ Factor}

\par 
The factor $0 < \eta \leq 1$ controls the maximum saturation of the input spike trains. By dialing $\eta$ closer and closer to zero, we may force the input spike trains have arbitrarily low saturation. However, note that, if $\rho\left(\text{abs}\left(A\right)\right) \geq 1$, we are not guaranteed that the recurrently circulating spikes will have low saturation, for if this condition is true, then the twice-as-large nonnegative LDS of the spiking implementation is not asymptotically stable, and thus we are not guaranteed that bounded inputs produce bounded states. For all of our numerical experiments, the condition $\rho\left(\text{abs}\left(A\right)\right) \geq 1$ happens to be true, however we retain the $\eta$ optimization since we found that it helps mitigate spike overflow nonetheless.

\subsection{Random generation of linear dynamical systems}

\par 
We used randomly generated linear dynamical systems with random inputs to compare the sample and theoretical error covariances. The role of randomness was merely to generate LDSs and inputs in an arbitrarily configurable manner -- the specific distribution from which these were sampled is not important, though we describe it here for completeness. To generate an appropriately normalized (LDS, input) pair with state dimension $m$, input dimension $n$, dynamics matrix spectral radius $\rho_0$, and $T$ time-steps, we first sampled the elements of the dynamics matrix $A$ independently and uniformly over the interval $[0.1, 1]$ and negated each off-diagonal element with probability $\frac{1}{2}$. The interval over which the matrix elements were sampled prevented any of them from being very close to zero, while the nonnegative diagonal elements resulted in smooth state sequences. To fix the spectral radius of $A$ at $\rho_0$, we scaled $A$ by $\frac{\rho_0}{\rho(A)}$. The input matrix $B$ was generated in the same manner as $A$, except all elements, including on-diagonals, were negated with probability $\frac{1}{2}$. Each of the $n$ components of the input was an integer-quantized sine wave lasting $T$ time-steps with amplitude $\eta p \ell$, random phase ($0$ or $\pi$, each with probability $\frac{1}{2}$), and uniformly sampled frequency. The initial state was fixed at zero to agree with the spiking LDS. To normalize the state sequence to $[-\eta p \ell, \eta p \ell]$, we drove the LDS using the generated inputs and computed the maximum absolute value of any state component across all time-steps. We then divided the input matrix $B$ by this value times $\eta p \ell$. Since the initial state was zero, this normalized the state sequence to $[-\eta p \ell, \eta p \ell]$. 

\par 
Using the above procedure, we generated a random LDS with $n = m = 5$, $\rho_0 = 0.9$, and $T = 2,400$. For the spiking LDS, we used frame length $\ell = 25$ which translates to 25 ms frames on TrueNorth, so the whole input spike sequence, consisting of 2,400 frames, lasted 60s. We used spike train dimension $p = 21$, the largest allowable value of this parameter on TrueNorth, and $\eta = 0.9$. The associated experimental results are shown in the main text in Fig. 2.

\par 
Next, we independently varied the input dimension $n$, the recurrent strength of the dynamics matrix, and the frame length $\ell$. In all three cases, we modified the LDS generation explained above to generate a sequence of LDSs and inputs. To vary the input dimension, we generated a sequence of LDSs with the same random dynamics matrix but with input dimensions ranging from 5 to 32 with linear spacing. Thus, each LDS had a different input matrix and different inputs. The spectral radius of the dynamics matrix was fixed to 0.9. To vary the recurrent strength of the dynamics matrix, we generated a sequence of LDSs whose dynamics matrices were scaled versions of the same matrix. By doing a fine-grained scan of this scale factor which took the spectral radius from 0.3 to 0.9, we stopped at 10 values of the scale factor corresponding to linearly spaced value of the recurrent strength ranging from 4.9 to 16.1. The same inputs were used for each LDS. The input matrix was also shared by all LDSs in the sequence, though it was scaled in each instance to confine the states to $[-\eta p \ell, \eta p \ell]$. Finally, varying the frame length is straightforward. The associated experimental results are shown in the main text in Fig. 3.

\section{Kalman filter}
\label{sec:appdxkf}

\par 
In wide-ranging domains, including engineering, econometrics, and neuroscience, the following problem is of great interest: Given a stream of noisy measurements of some system which is evolving according to stochastic dynamics, what is the best estimate of the current state of the system? Under certain modeling assumptions regarding the dynamics of the system and the nature of measurement, the optimal state estimate may be computed using a simple and intuitive algorithm called the Kalman filter, which processes the measurement stream in real-time and outputs the optimal state estimate at each time-step. What's more, in the limit where the Kalman filter has been running for many time-steps, the algorithm takes the form of a LDS. This means that we can leverage the spiking LDS described in the previous section to perform Kalman filtering. In Section \ref{subsec:kfoverview}, we provide a self-contained introduction to the Kalman filter. In Section \ref{subsec:sskf}, we describe the steady-state limit of the Kalman filter.

\subsection{Kalman filter overview}
\label{subsec:kfoverview}

\par
To make the state estimation problem described above mathematically well-defined, we must establish 1) a probabilistic dynamical model governing the system in question, 2) a probabilistic description of the measurement process, and 3) the definition of the ``optimal state estimate.''
\par 
Let $\bo{x}_t \in \mathbb{R}^m$ denote the state of the system at time-step $t$. We take the the initial state $\bo{x}_0$ to be a Gaussian random variable, $\bo{x}_0 \sim \mathcal{N}(\hat{\bo{x}}_0, P_0)$, and assume that the dynamics of the system are linear and Markovian with additive zero-mean Gaussian noise introduced at each time-step. Let $\bo{y}_t \in \mathbb{R}^n$ denote the measurement of the system at time-step $t$. We assume that a measurement is a linear function of the state at the same time-step, with additive zero-mean Gaussian noise. Altogether,
\begin{subequations}
\begin{align}
\bo{x}_t = \Phi\bo{x}_{t-1} + \bo{w}_t,& \:\:\: \bo{w}_t \sim \mathcal{N}(0, Q), \\
\bo{y}_t = H\bo{x}_t + \bo{v}_t,& \:\:\: \bo{v}_t \sim \mathcal{N}(0, R).
\end{align}
\label{eq:kfmodel}
\end{subequations}
We assume that the variables $\{ \hat{\bo{x}}_0, \bo{w}_1,\bo{w}_2, \ldots, \bo{v}_1, \bo{v}_2, \ldots \}$ are mutually independent. The posterior distribution
\begin{equation}
P(\bo{x}_t \: | \: \bo{y}_1, \ldots, \bo{y}_t )
\label{eq:post}
\end{equation}
encodes all there is to know about the state at time-step $t$ given the measurements at time-steps up to and including $t$. It is easy to show that the posterior is Gaussian and therefore fully characterized by its mean, denoted $\hat{\bo{x}_t}$, and its covariance, denoted $P_t$. The mean $\hat{\bo{x}}_t$ is the ``optimal state estimate" which we aim to compute. Suppose we knew $\hat{\bo{x}}_{t-1}$ and $P_{t-1}$ and wanted to know $\hat{\bo{x}}_t$ and $P_t$. As a first step, we could compute the mean and covariance of $P(\bo{x}_{t} \: | \: \bo{y}_1, \ldots, \bo{y}_{t-1} )$, i.e. the state mean and covariance obtained by propagating $\hat{\bo{x}}_{t-1}$ and $P_{t-1}$ to the subsequent time-step without incorporating knowledge of the measurement $\bo{y}_t$. We denote the propagated mean and covariance using $``-"$ superscripts:
\begin{subequations}
\label{eq:kfprop}
\begin{align}
\hat{\bo{x}}_t^- &= \Phi\hat{\bo{x}}_{t-1}, \label{eq:kfpropmean} \\
P_t^- &= \Phi P_{t-1} \Phi^T + Q \label{eq:kfpropcov}.
\end{align}
\end{subequations}
Based on the propagated state mean (Eq. \ref{eq:kfpropmean}), we predict that the measurement at time-step $t$ should be $H \hat{\bo{x}}_t^-$. However, at time-step $t$, we know the actual value of this measurement, namely, $\bo{y}_t$. Thus, the difference $\bo{y}_t - H \hat{\bo{x}}_t^-$, called the ``innovation,'' reflects the quality of the state estimate obtained by propagating from the previous time-step. If the innovation is zero, we should not modify our state estimate, but if it is nonzero, we should somehow modify the estimate to correct for the discrepancy. For this purpose, we define the Kalman gain
\begin{equation}
\label{eq:gain}
K_t = P_t^- H^T \left( H P_t^- H^T + R\right)^{-1}.
\end{equation}
We then update the propagated state mean and covariance according to
\begin{subequations}
\label{eq:kfcorrect}
\begin{align}
\hat{\bo{x}}_t &= \hat{\bo{x}}_t^- + K_t\left(\bo{y}_t - H \hat{\bo{x}}_t^-\right), \label{eq:kfcorrectmean} \\
P_t &= \left(I - K_t H\right)P_t^-. \label{eq:kfcorrectcov}
\end{align}
\end{subequations}
If we iteratively perform the updates of equations \ref{eq:kfprop}, \ref{eq:gain}, and \ref{eq:kfcorrect} with initial conditions $\hat{\bo{x}}_0$ and $P_0$, we correctly compute the posterior distribution (Eq. \ref{eq:post}) for all $t$. This algorithm is called the Kalman filter \cite{chui2017kalman}. 

\subsection{Steady-state Kalman filter}
\label{subsec:sskf}

\par 
Let us combine equations \ref{eq:kfpropcov}, \ref{eq:gain}, and \ref{eq:kfcorrectcov} to solve for the evolution of $P_t^-$ under the Kalman filter. We find:
\begin{equation}
\label{eq:dare}
P_t^- = \Phi \left(P_{t-1}^- -   P_{t-1}^- H^T \left( H P_{t-1}^- H^T + R\right)^{-1}  HP_{t-1}^- \right) \Phi^T + Q 
\end{equation}
This expression, known as a discrete-time algebraic Riccati equation, underscores the somewhat surprising fact that the evolution of the state covariance depends only on the system matrices and $t$, not on the measurement stream. It can be shown that if $\rho\left(\Phi \right) < 1$, then $P_t^-$ converges to a steady-state value $P_\text{SS}^-$ exponentially quickly as $t \rightarrow \infty$. This steady-state value is a fixed point of Eq. \ref{eq:dare}. Using $P_\text{SS}^-$ we may compute the steady-state Kalman gain
\begin{equation}
K_\text{SS} = P_\text{SS}^- H^T \left( H P_\text{SS}^- H^T + R\right)^{-1}.
\end{equation}
Combining this with equations \ref{eq:kfpropmean} and \ref{eq:kfcorrectmean}, we see that $\hat{\bo{x}}_t$ evolves, in the steady-state limit of the Kalman filter, according to
\begin{equation}
\hat{\bo{x}}_t = A_\text{SSKF} \hat{\bo{x}}_{t-1} + B_\text{SSKF} \bo{y}_t
\end{equation}
where $ A_\text{SSKF}$ and $B_\text{SSKF}$ are given by
\begin{equation}
A_\text{SSKF} = \Phi - K_\text{SS} H \Phi, \:\:\:  B_\text{SSKF} = K_\text{SS}
\end{equation}
and $A_\text{SSKF}$ satisfies $\rho\left(A_\text{SSKF} \right) < 1$. To summarize: assuming that the system whose state we wish to estimate is asymptotically stable, then the optimal Bayesian filter converges exponentially quickly to an asymptotically stable LDS, driven by the noisy measurements, whose state corresponds to the optimal state estimate \cite{chui2017kalman}. By mapping this LDS onto a spiking architecture, we may run a spiking version of the Kalman filter.

\section{ECoG Kalman filter fit}
\label{sec:appdxecog}

\par 
For each of the 38 trials, we fit the KF model (Eq. \ref{eq:kfmodel}) to the data from all the other trials (i.e. leave-one-out cross validation) using standard maximum likelihood techniques. We imposed the constraint that the dynamics matrix $\Phi$ satisfied
\begin{equation}
\Phi = \thbthmat{1}{\Delta t}{0}{0}{\phi}{0}{0}{0}{1}
\end{equation}
where $\phi$ is a free parameter which determines the velocity dynamics. Based on the parameters obtained from the fitting procedure, we constructed the steady-state Kalman filter matrices $A_\text{SSKF}$ and $B_\text{SSKF}$. These matrices have the form
\begin{equation}
A_\text{SSKF} = \tbtmat{A'_\text{SSKF}}{\bo{a}_\text{bias}}{0}{1},\:\:\:
B_\text{SSKF} = \tbomat{B'_\text{SSKF}}{0}
\end{equation}
where $A'_\text{SSKF}$ is a $2 \times 2$ matrix, $\bo{a}_\text{bias}$ is a 2-dimensional vector, and $B'_\text{SSKF}$ is a $2 \times N_\text{elecs}$ matrix, where $N_\text{elecs} = 73$. Note that it is important that the initial state of this system has its last component equal to one, the constant bias component. This is not possible in the spiking implementation, which requires that the initial state is zero. However, we may construct an equivalent system by moving the constant bias component from the state to the inputs, and using the system matrices $A'_\text{SSKF}$ and $\left( \begin{array}{cc} B'_\text{SSKF} & \bo{a}_\text{bias} \end{array} \right)$.


%% file: nips_2018.bbl
\begin{thebibliography}{10}

\bibitem{mead}
C.~Mead.
\newblock Neuromorphic electronic systems.
\newblock {\em Proceedings of the IEEE}, 78(10):1629--1636, Oct 1990.

\bibitem{Esser201604850}
Steven~K. Esser, Paul~A. Merolla, John~V. Arthur, Andrew~S. Cassidy,
  Rathinakumar Appuswamy, Alexander Andreopoulos, David~J. Berg, Jeffrey~L.
  McKinstry, Timothy Melano, Davis~R. Barch, Carmelo di~Nolfo, Pallab Datta,
  Arnon Amir, Brian Taba, Myron~D. Flickner, and Dharmendra~S. Modha.
\newblock Convolutional networks for fast, energy-efficient neuromorphic
  computing.
\newblock {\em Proceedings of the National Academy of Sciences}, 2016.

\bibitem{rueckauer2017conversion}
Bodo Rueckauer, Yuhuang Hu, Iulia-Alexandra Lungu, Michael Pfeiffer, and
  Shih-Chii Liu.
\newblock Conversion of continuous-valued deep networks to efficient
  event-driven networks for image classification.
\newblock {\em Frontiers in neuroscience}, 11:682, 2017.

\bibitem{NIPS20155862}
Steve~K Esser, Rathinakumar Appuswamy, Paul Merolla, John~V. Arthur, and
  Dharmendra~S Modha.
\newblock Backpropagation for energy-efficient neuromorphic computing.
\newblock In C.~Cortes, N.~D. Lawrence, D.~D. Lee, M.~Sugiyama, and R.~Garnett,
  editors, {\em Advances in Neural Information Processing Systems 28}, pages
  1117--1125. Curran Associates, Inc., 2015.

\bibitem{hunsberger2016training}
Eric Hunsberger and Chris Eliasmith.
\newblock Training spiking deep networks for neuromorphic hardware.
\newblock {\em arXiv preprint arXiv:1611.05141}, 2016.

\bibitem{zenke2017superspike}
Friedemann Zenke and Surya Ganguli.
\newblock Superspike: Supervised learning in multi-layer spiking neural
  networks.
\newblock {\em arXiv preprint arXiv:1705.11146}, 2017.

\bibitem{schuman2017survey}
Catherine~D Schuman, Thomas~E Potok, Robert~M Patton, J~Douglas Birdwell,
  Mark~E Dean, Garrett~S Rose, and James~S Plank.
\newblock A survey of neuromorphic computing and neural networks in hardware.
\newblock {\em arXiv preprint arXiv:1705.06963}, 2017.

\bibitem{furber2014spinnaker}
Steve~B Furber, Francesco Galluppi, Steve Temple, and Luis~A Plana.
\newblock The {SpiNNaker} project.
\newblock {\em Proceedings of the IEEE}, 102(5):652--665, 2014.

\bibitem{merolla2014million}
Paul~A Merolla, John~V Arthur, Rodrigo Alvarez-Icaza, Andrew~S Cassidy, Jun
  Sawada, Filipp Akopyan, Bryan~L Jackson, Nabil Imam, Chen Guo, Yutaka
  Nakamura, et~al.
\newblock A million spiking-neuron integrated circuit with a scalable
  communication network and interface.
\newblock {\em Science}, 345(6197):668--673, 2014.

\bibitem{carney2017neuromorphic}
R~Carney, K~Bouchard, P~Calafiura, D~Clark, D~Donofrio, M~Garcia-Sciveres, and
  J~Livezey.
\newblock Neuromorphic {Kalman} filter implementation in {IBM’s TrueNorth}.
\newblock In {\em Journal of Physics: Conference Series}, volume 898, page
  042021. IOP Publishing, 2017.

\bibitem{kalman1960new}
Rudolph~Emil Kalman.
\newblock A new approach to linear filtering and prediction problems.
\newblock {\em Journal of basic Engineering}, 82(1):35--45, 1960.

\bibitem{malik2011efficient}
Wasim~Q Malik, Wilson Truccolo, Emery~N Brown, and Leigh~R Hochberg.
\newblock Efficient decoding with steady-state kalman filter in neural
  interface systems.
\newblock {\em IEEE Transactions on Neural Systems and Rehabilitation
  Engineering}, 19(1):25--34, 2011.

\bibitem{gilja2012high}
Vikash Gilja, Paul Nuyujukian, Cindy~A Chestek, John~P Cunningham, M~Yu Byron,
  Joline~M Fan, Mark~M Churchland, Matthew~T Kaufman, Jonathan~C Kao, Stephen~I
  Ryu, et~al.
\newblock A high-performance neural prosthesis enabled by control algorithm
  design.
\newblock {\em Nature neuroscience}, 15(12):1752, 2012.

\bibitem{sussillo2012recurrent}
David Sussillo, Paul Nuyujukian, Joline~M Fan, Jonathan~C Kao, Sergey~D
  Stavisky, Stephen Ryu, and Krishna Shenoy.
\newblock A recurrent neural network for closed-loop intracortical
  brain--machine interface decoders.
\newblock {\em Journal of neural engineering}, 9(2):026027, 2012.

\bibitem{dethier2013design}
Julie Dethier, Paul Nuyujukian, Stephen~I Ryu, Krishna~V Shenoy, and Kwabena
  Boahen.
\newblock Design and validation of a real-time spiking-neural-network decoder
  for brain--machine interfaces.
\newblock {\em Journal of neural engineering}, 10(3):036008, 2013.

\bibitem{sawada2016truenorth}
Jun Sawada, Filipp Akopyan, Andrew~S Cassidy, Brian Taba, Michael~V Debole,
  Pallab Datta, Rodrigo Alvarez-Icaza, Arnon Amir, John~V Arthur, Alexander
  Andreopoulos, et~al.
\newblock Truenorth ecosystem for brain-inspired computing: scalable systems,
  software, and applications.
\newblock In {\em High Performance Computing, Networking, Storage and Analysis,
  SC16: International Conference for}, pages 130--141. IEEE, 2016.

\bibitem{dichter2018control}
Benjamin~K Dichter, Jonathan~D Breshears, Matthew~K Leonard, and Edward~F
  Chang.
\newblock The control of vocal pitch in the human laryngeal motor cortex.
\newblock {\em Cell (in press)}, 2018.

\bibitem{bouchard2013functional}
Kristofer~E Bouchard, Nima Mesgarani, Keith Johnson, and Edward~F Chang.
\newblock Functional organization of human sensorimotor cortex for speech
  articulation.
\newblock {\em Nature}, 495(7441):327, 2013.

\bibitem{chui2017kalman}
Charles~K Chui, Guanrong Chen, et~al.
\newblock {\em Kalman filtering}.
\newblock Springer, 2017.

\bibitem{NIPS2009_3665}
Robert Wilson and Leif Finkel.
\newblock A neural implementation of the kalman filter.
\newblock In Y.~Bengio, D.~Schuurmans, J.~D. Lafferty, C.~K.~I. Williams, and
  A.~Culotta, editors, {\em Advances in Neural Information Processing Systems
  22}, pages 2062--2070. Curran Associates, Inc., 2009.

\bibitem{eliasmith2004neural}
Chris Eliasmith and Charles~H Anderson.
\newblock {\em Neural engineering: Computation, representation, and dynamics in
  neurobiological systems}.
\newblock MIT press, 2004.

\bibitem{sussillo2009generating}
David Sussillo and Larry~F Abbott.
\newblock Generating coherent patterns of activity from chaotic neural
  networks.
\newblock {\em Neuron}, 63(4):544--557, 2009.

\bibitem{nicola2017supervised}
Wilten Nicola and Claudia Clopath.
\newblock Supervised learning in spiking neural networks with force training.
\newblock {\em Nature communications}, 8(1):2208, 2017.

\bibitem{depasquale2018full}
Brian DePasquale, Christopher~J Cueva, Kanaka Rajan, LF~Abbott, et~al.
\newblock full-force: A target-based method for training recurrent networks.
\newblock {\em PloS one}, 13(2):e0191527, 2018.

\bibitem{Trautmann229252}
Eric Trautmann, Sergey Stavisky, Subhaneil Lahiri, Katherine Ames, Matthew
  Kaufman, Stephen Ryu, Surya Ganguli, and Krishna Shenoy.
\newblock Accurate estimation of neural population dynamics without spike
  sorting.
\newblock {\em bioRxiv}, 2017.

\bibitem{churchland2012neural}
Mark~M Churchland, John~P Cunningham, Matthew~T Kaufman, Justin~D Foster, Paul
  Nuyujukian, Stephen~I Ryu, and Krishna~V Shenoy.
\newblock Neural population dynamics during reaching.
\newblock {\em Nature}, 487(7405):51, 2012.

\bibitem{shenoy2013cortical}
Krishna~V Shenoy, Maneesh Sahani, and Mark~M Churchland.
\newblock Cortical control of arm movements: a dynamical systems perspective.
\newblock {\em Annual review of neuroscience}, 36, 2013.

\end{thebibliography}
